\documentclass[manuscript,screen,nonacm]{acmart}

\AtBeginDocument{%
  \providecommand\BibTeX{{%
    \normalfont B\kern-0.5em{\scshape i\kern-0.25em b}\kern-0.8em\TeX}}}

\copyrightyear{2021}
\acmYear{2021}
\acmJournal{THRI}

\usepackage{wrapfig}
\usepackage{tikz}                                                          
\usepackage{graphicx} 
\usepackage{amsthm}
\usepackage{amsmath}
 
\usepackage{epstopdf}
\usepackage{graphics}
\numberwithin{equation}{section}
\usepackage{color}      
\usepackage{hyperref}   
\usepackage{verbatim}   
\usepackage{xspace}
\usepackage{float}
\usepackage{csquotes}
\usepackage{footnote}
\makesavenoteenv{tabular}
\makesavenoteenv{table}
\usepackage[]{threeparttable}
\usepackage{enumitem}
\usepackage{multicol}
\usepackage{caption}
\usepackage{subcaption}
\newcommand{\xxnote}[3]{}
\ifx\hidenotes\undefined
  \usepackage{color}
  \renewcommand{\xxnote}[3]{\color{#2}{#1: #3}}
\fi

\usepackage{macros}

\usepackage{graphicx}
\usepackage{hyperref}
\usepackage{cleveref}
\usepackage{caption} 
\usepackage{keyval}
\usepackage[ruled,vlined]{algorithm2e}
\usepackage{color}

\newcommand{\idest}{{\it i.e.}, }
\newcommand{\exempli}{{\it e.g.}, }

\newtheorem{definition}{\textbf{Definition}}

\usepackage{csvsimple}
\usepackage{siunitx}
\usepackage{soul}
\usepackage{multirow}

\usepackage{titlesec}

\begin{document}






\title{Core Challenges of Social Robot Navigation: A Survey}

\author{Christoforos Mavrogiannis}
\affiliation{
  \institution{Paul G. Allen School of Computer Science \& Engineering, University of Washington}
  \country{USA}
  \orcid{0000-0003-4476-1920}
}
\email{cmavro@cs.washington.edu}

\author{Francesca Baldini}
\affiliation{%
  \institution{Honda Research Institute and California Institute of Technology}
  \country{USA}
  }
\email{fbaldini@caltech.edu}

\author{Allan Wang}
\affiliation{%
  \institution{Robotics Institute, Carnegie Mellon University}
  \country{USA}}
\email{allanwan@andrew.cmu.edu}

\author{Dapeng Zhao}
\affiliation{%
  \institution{Robotics Institute, Carnegie Mellon University}
  \country{USA}}
\email{dapengz@andrew.cmu.edu}

\author{Pete Trautman}
\affiliation{%
  \institution{Honda Research Institute}
  \country{USA}}
\email{peter.trautman@gmail.com}

\author{Aaron Steinfeld}
\affiliation{%
  \institution{Robotics Institute, Carnegie Mellon University}
  \country{USA}}
\email{steinfeld@cmu.edu}

\author{Jean Oh}
\affiliation{%
  \institution{Robotics Institute, Carnegie Mellon University}
  \country{USA}}
\email{jeanoh@cmu.edu}

\renewcommand{\shortauthors}{C. Mavrogiannis, et al.}

\begin{abstract}
Robot navigation in crowded public spaces is a complex task that requires addressing a variety of engineering and human factors challenges. These challenges have motivated a great amount of research resulting in important developments for the fields of robotics and human-robot interaction over the past three decades. Despite the significant progress and the massive recent interest, we observe a number of significant remaining challenges that prohibit the seamless deployment of autonomous robots in public pedestrian environments. In this survey article, we organize existing challenges into a set of categories related to broader open problems in motion planning, behavior design, and evaluation methodologies. Within these categories, we review past work, and offer directions for future research. Our work builds upon and extends earlier survey efforts by a) taking a critical perspective and diagnosing fundamental limitations of adopted practices in the field and b) offering constructive feedback and ideas that we aspire will drive research in the field over the coming decade.
\end{abstract}


\maketitle

\section{Introduction}

The vision of mobile robots navigating seamlessly through public spaces has inspired a significant amount of research over the past three decades. While early efforts were driven by engineering principles, it quickly became evident that the problem is inherently interdisciplinary, requiring insights from fields such as human-robot interaction, psychology, design research, and sociology. This understanding resulted in the emergence of a dedicated field of research, which is commonly referred to as \emph{social robot navigation}, broadly addressing all algorithmic, behavioral, evaluation, and design aspects of the problem.

 
Early work focused on the development of tour guide robots. The RHINO~\citep{rhino-robot} and MINERVA~\citep{thrun_ijrr2000} robots, deployed in Museums in Germany and USA in the late 1990's, exemplified the vision of integrating robots in human environments. Leveraging state-of-the-art probabilistic methods for localization and mapping \citep{coastal-navigation,monte-carlo-localization,self-calibrate-minerva}, both robots successfully served thousands of tour guide requests under unscripted interaction settings with human visitors, pioneering the nascent field of human-robot interaction in natural spaces \citep{short-term-interaction,interaction-minerva}. Despite their pioneering results, the RHINO and MINERVA experiments uncovered a series of fundamental engineering and interaction challenges underlying the deployment of autonomous robots in human populated spaces. 

Over the 20 years that followed the RHINO and MINERVA studies, the field of social robot navigation has seen many exciting developments. Recently, the interest in the field has been propelled even further as a result of the overlap of social navigation in neighboring fields such as autonomous driving. Further, recent and ongoing large projects such as SPENCER~\citep{spencer-eu} or CROWDBOT~\citep{crowdbot-eu} funded by the European Union are indicative of the momentum and growth of the field. In this review, we broadly organize the literature into a set of core challenges that represent major classes of ongoing research and open problems. In particular, we focus on:
\begin{itemize}
    \item Planning challenges: \emph{How should a robot plan its motion through a crowd of navigating humans to ensure safety and efficiency?}
    \item Behavioral challenges: \emph{What types of social signals can be inferred from human behavior and what types of behaviors should a robot exhibit to ensure social norm compliance?}
    \item Evaluation challenges: \emph{How can we evaluate a social robot navigation system?}
\end{itemize}
For each class of challenges, we review relevant literature, diagnose problems, and offer research directions to address these problems. It should be noted that our perspective is driven by computational insights. As such, the proposed categorization, the selection of the literature, and the discussion is done from a computational lense. While this allows us to cover a wide range of important topics, there are additional challenges from the broader space of Robotics and Human-Robot Interaction (HRI) that are outside of the scope of this review, \exempli robot design or perception pipelines.

\subsection{Contributions}

Overall, this article contributes the following:

\begin{itemize}
    \item A formal definition of social robot navigation that encapsulates important pieces of the problem. This definition is motivated by past and ongoing efforts in the field.
    
    \item A holistic discussion of the literature in the area of social robot navigation spanning three decades of work. This discussion features a taxonomy of past and active trends.

    \item An enumeration of fundamental challenges in social robot navigation, representing our perspective on the most important open problems that require further research. 

    \item A discussion of possible research directions that could address the challenges identified.

\end{itemize}

This article builds upon and complements past surveys on the topic and related areas~\citep{kruse13-navSurvey,riosmartinez-survey,thomaz2016computational,charalampous17,rudenko2019-predSurvey}. In particular, the survey of \citet{riosmartinez-survey} specifically focuses on the sociological aspects underlying the development of social robot navigation frameworks. \citet{charalampous17} focuses on the perception challenges arising in social-navigation research, whereas in the survey of \citet{thomaz2016computational} the focus is on computational human-robot interaction more broadly. Finally, the survey of \citet{rudenko2019-predSurvey} specifically focuses on trajectory prediction and not necessarily on the challenges underlying embodiment on robot platforms or robot navigation.

Our article is inspired by and close in principle with the survey of \citet{kruse13-navSurvey} in that it attempts a more holistic treatment of the problem. In fact, we are following up on the conclusions of \citet{kruse13-navSurvey}, calling for a ``holistic theory of human-aware navigation''. While we are still far from a unified theory on the topic, we offer a unique bottom-up perspective, starting from working problem definitions, and following up with a taxonomization of past work, a layout of open problems and a selection of proposed directions that we aspire to be a valuable resource for researchers and practitioners over the next decade. 

This article is organized as follows: Sec.~\ref{sec:problem} offers working definitions for navigation and social robot navigation to help scope the survey; Sec.~\ref{sec:plan_challenge} delves into the planning problems uncovered in social robot navigation applications, highlighting the coupling between prediction and planning in crowded human environments; Sec.~\ref{sec:behavioral} focuses on the behavioral aspects underlying the motion of robots in public spaces, ranging from the theory of proxemics to the importance of intention and formation recognition; Sec.~\ref{sec:evalchal} discusses methodologies of evaluation and points out fundamental issues in simulation and experimental validation; and finally, Sec.~\ref{sec:conclusion} summarizes our key insights and enlists remaining open questions.


\section{Problem Formulation}\label{sec:problem}



Navigation is the task of following a collision-free, efficient route from an initial location to a destination. 
In general, navigation has been studied as a hierarchical planning problem that is composed of \textit{global planning} and \textit{local planning}. A global planner is designed to find a sequence of waypoints to reach a destination; in general, a global planner assumes a static environment where traversability depends mainly on the environmental context such as terrain types or obstacles. Given a route or a set of waypoints found by a global planner, a local planner aims to safely navigate to a next waypoint; thus, the primary objective of local planning is to avoid collision while satisfying various types of run-time, dynamic constraints as well as vehicle constraints. In this context, social robot navigation can be seen as local planning specifically focused on a dynamic environment where moving obstacles are present such as pedestrians.

In this article, we define social navigation as a navigation task in a dynamic human environment where each agent has both private (hidden) and public (visible) objectives, that is, to efficiently reach a goal while abiding by social rules/norms. Although there is the notion of a loosely defined group, \exempli a crowd, a solution to social navigation is generally sought in a distributed fashion where every agent in an environment makes decisions more or less independently. Explicit coordination between agents is rare in a sparse setting but can be found commonly in a dense crowd, \exempli gesturing for someone to pass, a dyad holding hands, etc. 


We use the following formal definitions to broadly scope the target problem of this article.

\begin{definition}[Navigation] 
Consider a planar workspace $\mathcal{C}\subseteq \mathbb{R}^2$ and denote by $\mathcal{C}_{obs}\in\mathcal{C}$ the subspace of $\mathcal{C}$ that is occupied by static obstacles. Assume that an agent\footnote{By agent, we refer to an \emph{entity} (individual or group).} $a$ is embedded in $\mathcal{C}$, lying at a configuration $q\in\mathcal{Q}\subseteq SE(2)$ and occupying a finite area $A(q)\subseteq\mathcal{C}$. Starting from $q$, the agent intends to reach a goal $g\in\mathcal{Q}$ while avoiding collisions with the static environment. For agent $a$, navigation is the task of following a collision free path $\tau:[0,1]\to\mathcal{Q}$ connecting $q$ to $g$. To this end, agent $a$ is solving a problem of the form:
\begin{equation}
\begin{aligned}
\tau = \arg\min_{\tau\in\mathcal{T}} \quad & c(\tau)\\
\textrm{s.t.} \quad & A\left(\tau(a)\right)\notin\mathcal{C}_{obs}, \quad t\in [0,1]\\
\quad & \tau(0) = q\\
\quad & \tau(1) = g\\
\end{aligned}\mbox{,}\label{eq:nav}
\end{equation}
where $t$ is a normalized notion of time, $\mathcal{T}$ is a space of paths, and $c:\mathcal{T}\to\mathbb{R}$ is a cost function describing additional path specifications (\exempli time to goal or path smoothness) and taking into account environmental properties (\exempli terrain traversability). 
\end{definition}


In social navigation, multiple mobile agents share an environment where each agent is solving one's own navigation problem. Here, an agent can observe other agents' states but can only control its own actions. Because every agent takes others into consideration during their decision-making process, agents influence one another in indirect, subtle ways. In addition to an agent's \textit{self-interested} objective represented by a  private cost function, navigation among pedestrians imposes a \textit{public-interested} objective that depends on a social context and the interactions among the agents. 

We assume that agents are not explicitly communicating with each other, \idest an agent does not have access to the goal or path of  other agents in an environment. The uncertainty induced by the lack of knowledge of such parameters introduces the need for the incorporation of prediction mechanisms over the behaviors of others but also the need for frequent replanning to ensure adaptation to the dynamic environment. 

\begin{definition}[Social Navigation] 
Consider a set of $n_{t}> 1$ agents, $a_{i}$, $i\in\mathcal{N}=\{1,\dots, n_{t}\}$, lying at configurations $q_{i}\in\mathcal{Q}$ with volumes $A_{i}(q_{i})$ at some time $t\geq0$. Agents intend to reach their individual destinations $g_{i}$ while avoiding collisions with the static environment and abiding by social norms (\exempli respecting the personal space of others). 
At planning time, agent $i$ generates a path $\tau_{i}:[0,1]\to\mathcal{Q}$, extracted by solving an optimization problem of the form:
\begin{equation}
\begin{aligned}
\tau_{i} = \arg\min_{\tau\in\mathcal{T}} \quad & c_{i}(\tau_{i}) + \lambda_{i} c^{s}_{i}(\tau_{i}, \tilde{\tau}_{-i})\\
\textrm{s.t.} \quad & A_{i}\left(\tau_{i}(a)\right)\notin\mathcal{C}_{obs}, \forall t\in [0,1]\\
\quad & A_{i}\left(\tau_{i}(a)\right) \cap A_{j}\left(\tilde{\tau}_{j}(a)\right) = \emptyset,\\ \quad & \qquad \qquad \forall t\in [0,1], j\neq i\\
\quad & \tau(0) = q_{i}\\
\quad & \tau(1) = g_{i}\\
\end{aligned}\label{eq:socialnav}
\end{equation}
where $c_{i}:\mathcal{T}\to\mathbb{R}$ is an individual cost corresponding to individual path specifications; $c^{s}_{i}:\mathcal{T}^2\to\mathbb{R}$, a cost\footnote{Note that this cost formulation is generic enough to possibly capture a wide variety of behaviors, ranging from social compliance to adversarial behavior. Rather than defining an explicit social cost, which is the subject of ongoing research, we define an abstract term to capture the space of possible objectives that could fall under the category of a cost that accounts for multiagent interaction.} describing social considerations, such as personal space, 
taking into account predictions\footnote{These can be explicit path predictions in the form of paths or implicit predictions encoding an abstraction over the behavior of other agents.} about the future behavior of others $\tilde{\tau}_{-i} = (\tilde{\tau}_{2},\dots, \tilde{\tau}_{n_{t}})$; and $\lambda_{i}$ a weight. We assume that agents do not have access to the complete navigation parameters of others. Specifically, we assume that at planning time agent $i$ is not aware of the navigation costs $c_{j}$, $c_{j}^{s}$ and weight $\lambda_{j}$, $\tau_{j}$. We refer to this particular instance of navigation as social navigation.
\end{definition}

\section{Core Planning Challenges of Social Navigation} \label{sec:plan_challenge}

\begin{figure*}
      \centering
\includegraphics[width=1\linewidth]{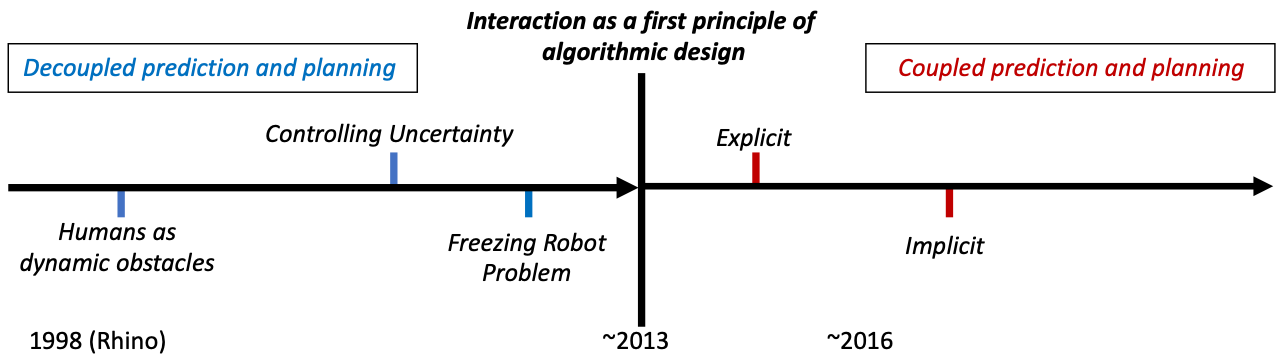}
      \caption{Timeline of trends in social robot navigation research. Most of these threads are still active to some extent.}
      \label{fig:timeline}
\end{figure*}

We broadly partition the literature into two classes: \emph{decoupled prediction and (navigation) planning} and \emph{coupled prediction and planning}; we separate decoupled and coupled approaches based on the recognition that interaction is a first principle of social navigation algorithmic design (see Figure~\ref{fig:timeline}).  In particular, we draw attention to the two different approaches: treating prediction and planning as independent or as coupled.  For the former, a prediction module feeds into a robot objective, but \emph{how} the agents will respond to robot action is not accounted for (often referred to as ``dynamic obstacle avoidance'').  For the latter, robot action and external agent prediction are treated holistically: agent prediction is a function of the decisions that the robot makes.  Coupling prediction and planning creates a broad theoretical umbrella: joint probability distributions, braid dynamics, interactive partially observable Markov decision processes (POMDPs), game-theoretic approaches, and coupled dynamical systems are all specific coupled prediction and planning formulations.

\subsection{Decoupled Prediction and Planning}

Decoupled approaches can be further classified into a set of dominant trends: a) planning while treating humans as dynamic, but non-responsive obstacles; b) planning using interaction-agnostic models of uncertainty; and c) planning by following hand-designed social objectives.

\subsubsection{Humans as Dynamic, Non-responsive Obstacles.\label{sec:movingobstacles}}

Early works were driven by system design for specific real-world applications, such as autonomous museum tour guidance. For instance, \emph{RHINO} \citep{rhino-robot} and \emph{MINERVA} \citep{thrun_ijrr2000} were some of the first known robotic systems to interact with human visitors in real-world museums. Deployed in Bonn, Germany, and Washington D.C., USA, in 1997 and 1998, respectively, these robots successfully served thousands of tour guide requests, and navigated alongside museum visitors over several weeks. The demonstrated capabilities of these systems motivated additional efforts in the tour-guide domain such as Robox~\citep{kai-robox}, Mobot~\citep{mobot}, Rackham~\citep{rackham}, and CiceRobot~\citep{cicerobot}. Finally, progress in the field was incentivized through competitions such as the AAAI Mobile Robot Challenge, organized by the American Association for Artificial Intelligence (AAAI)~\citep{maxwell2007,michaud2007}.

While these systems achieved important milestones in terms of autonomous navigation capabilities and serving guidance requests, they totally ignored the human component. In fact, their underlying navigation frameworks (\exempli \citep{Fox98ahybrid,Montemerlo}) treated humans as non-reactive obstacles without explicitly modeling human navigation, social engagement or interaction with and between humans. Abstracting away these processes has generally proved to provide a practical solution in terms of collision avoidance which has inspired a significant amount of research over the years \citep{bellotto,fokamuseum,park2012robot}. However, the lack of models for human behavior tends to result in subtle robot behaviors that negatively impact user experience. For instance, robots that treat humans as non-reactive obstacles (and thus do not make predictions about human motion) tend to hinder or block human paths \citep{ziebart2009} with their motion. An interesting phenomenon often observed is the so-called ``reciprocal dance" \citep{feurtey2000simulating}: the failure of the robot to predict the human's motion leads to robot motion that surprises the human, who in turn reacts unpredictably, contributing to a short oscillatory interaction. Such issues motivated the introduction of models for reasoning about uncertainty in crowd navigation.


\subsubsection{Modeling Uncertainty.}

The practical issues observed during the first wave of robot deployments in crowded human environments motivated a shift in the navigation architecture. In particular, the second wave of work in the area focused on the development of methodologies for reasoning about the inherent uncertainty of the domain. For instance, \citet{dutoit-tro} presented a receding-horizon control framework that incorporated predictive uncertainty in the robot's decision making. \citet{Thompson} introduced a probabilistic model of human motion based on individual human intent inference, designed to assist in motion planning problems. \citet{Joseph11} proposed a Bayesian framework for reasoning about individual human motion patterns to inform a motion prediction pipeline. Similarly, \citet{Bennewitz05} extracted patterns of human motion in a crowded environment and derive a hidden Markov model to perform online human motion prediction. \citet{vaibhav-co-nav} introduce a motion prediction framework that makes use of biomechanical features to anticipate human turning actions. 

Despite the principled introduction of models for reasoning about uncertainty, these works treat human agents as \emph{individual non-interactive entities}. The lack of explicit coupling over the possible motion of human agents often results in an \emph{uncertainty explosion} as observed by \citet{dutoit-tro}. In practice, this may cause the robot to overestimate the uncertainty over the unfolding crowd motion, and thus falsely infer that no collision-free paths exist. This phenomenon may effectively yield the \emph{freezing robot problem} as described by \citet{trautman-ijrr-2015}, who pointed out that a practical way to resolve it is to enable the robot to explicitly expect human cooperation.


\subsubsection{Engineering Social Awareness.}

A body of work focuses on generating motion that abides by a set of selected social norms or conventions towards enhancing the levels of comfort of human bystanders.  Some works, following the insights of proxemics theory \citep{hall-1966}, propose planning frameworks designed to generate motion that respects the personal space of co-navigating pedestrians. \citet{KnepperROMAN12} distill the concept of \emph{civil inattention} observed by \citet{goffman} into a a multirobot path planner. For instance, \citet{Sisbot07TRO} propose a motion planner that generates safe motion that occurs within the field of view of humans. A class of works extracts social features of human navigation from demonstrations. For example, \citet{learning-social-robot} learn a set of dynamic navigation prototypes and use them to design dynamic costmaps that capture objective and subjective human navigation objectives. The works of \citet{ziebart2009, Vasquez14, Kim2016, kuderer-ijrr-2016} employ inverse reinforcement learning as a technique to recover models of human navigation objectives and use them to design motion planners for humanlike robot navigation.





\subsection{Coupled Prediction and Planning}
Decoupled approaches tend to follow the conventional paradigm of decoupling motion prediction from motion planning, often resulting in undesired phenomena such as the freezing robot problem~\citep{trautman-ijrr-2015} or the reciprocal dance problem~\citep{feurtey2000simulating}. These observations motivated a new research trend focusing on modeling multiagent interaction in crowd domains as cooperative collision avoidance (CCA). This modeling paradigm is further backed by studies highlighting the cooperative nature of human navigation (see \exempli \citet{Wolfinger95}) but also by the provable hardness of inference over multiagent behavior (\exempli multistep probabilistic inference on Bayesian  networks  has  been  shown  to  be  NP-hard \citep{cooper90}).

Within this body of work, we observe two main trends depending on the type of cooperation models they employ: explicit approaches and implicit approaches.

\subsubsection{Explicit Approaches}

Explicit approaches leverage insights about the structure of multiagent collision avoidance to couple prediction and planning.

For instance, some works employ geometric or topological representations to model the coupling among the trajectories of multiple navigating agents. For instance, Mavrogiannis et al. employ topological representations such as braids \citep{birman,mavrogiannis_ijrr,MavBluKne_IROS2017}, and topological invariants \citep{Berger2001invariants,Mavrogiannis18,MavKne_WAFR_2018} to abstract away multiagent dynamics into a set of enumerable classes of joint behavior. Using these abstractions, they perform symbolic inference for multiagent motion prediction and contribute motion planners for legible and socially compliant motion generation.

Some works model multiagent collision avoidance in social robot navigation by leveraging tools from game theory. For instance, \citet{Turnwald2019} model the problem as a non-cooperative, static, multiplayer game and iteratively solve for Nash equilibria corresponding to collision-free strategies. \citet{fridovich2020ilqr} treat the problem as a multiplayer, general-sum differential game which they solve using an iterative approach resembling iterative linear quadratic regulator (ILQR). 

Other works leverage insights about the structure of selected real-world navigation tasks. For example,   \citet{Chatterjee1} amd \citet{wang-split-merge} propose a ``Gestalt'' approach using frameworks that cluster navigating pedestrians into groups. \citet{nanavati-dyad} focus on leader-follower tasks involving a guiding robot and a following human and extract an empirical model of human following behavior that enables a robot to plan effective maneuvers leading humans to their destinations.




\subsubsection{Implicit Approaches}

Implicit approaches leverage insights about the principles underlying cooperative collision avoidance but do not impose explicit constraints on its structure.

A significant amount of research focuses on learning navigation strategies by observing example trajectories from human or simulated datasets.  For instance, \citet{learning-social-robot} learn motion prototypes to guide navigation. A popular computational paradigm in this area is Inverse Reincorcement Learning (IRL). \citet{ziebart2009} propose an IRL framework that generates robot motion of minimal disruption to expected human paths in an obstacle-occupied environment. \citet{Henry-2010} extend the general IRL paradigm to more crowded dynamic domains. \citet{kuderer-ijrr-2016} incorporate a featurization that captures discrete (\exempli passing from the right or left side) and continuous features of cooperation (\exempli speed adjustment) into the general IRL paradigm. \citet{trautman-ijrr-2015} introduce \emph{interacting Gaussian processes}, a mechanism explicitly accounting for human cooperation for human motion prediction by introducing interdependencies across individual Gaussian processes tracking the motion of pedestrians. Their extensive real-world experiments yielded safe robot performance in densely crowded environments.  


A number of works have proposed deep reinforcement learning models for prediction in crowd navigation domains. For instance, \citet{chen17-sa-cadrl,unfrozen-unlost} employ CADRL \citep{chen17-cadrl}, a decentralized deep reinforcement learning framework for socially aware multiagent collision avoidance. \citet{everett18-lstm} relaxes prior assumptions with an actor-critic variant \citep{Babaeizadeh-GA3C}, simultaneously learning policies and agent motion models. Furthermore, \citet{social-nav-gan-end-to-end} employ generative adversarial imitation learning, trained on a dataset generated using the social force model~\citep{Helbing95}. Finally, \citet{chen19-crowdNav} employ attention-based reinforcement learning to produce interaction aware collision avoidance behaviors. 

Recently, several deep learning-based approaches have been proposed for crowd prediction where an agent's planning objective is not well defined~\citep{alahi16-socialLstm,gupta18-socialGan,vemula18-attention,mohamed20-stgcnn,zhao2020}. These models perform well in predicting the social interactions of a crowd; however, they are not suitable for navigation because they do not plan for a specific goal and also because the models are not trained to avoid collision, resulting in risky behaviors. To this end, \citet{tsai20} introduce a generative adversarial network architecture for social navigation where an agent is trained to navigate using multiple objectives including private intention and social compliance.




Finally, a class of works leverage intent-inference mechanisms for behavior prediction and interaction-aware navigation. Specifically, several works condition prediction and motion generation on estimates of other agents' goals acquired from past behavior observation. For instance, \citet{intention-aware-crowd} and \citep{intention-aware-momdp} reason about the uncertainty of human intent by employing \emph{partially observable Markov decision processes} (POMDP) and \emph{mixed observability Markov decision processes} (MOMDP) frameworks. Some other works employ motion indicators to improve intent inference, such as anticipatory biomechanical turn indicators \citep{vaibhav-co-nav} or libraries of motion level indicators \citep{fast-target-prediction}. Similarly, \citet{disney-robot} employ a data-driven model that enables a robot to avoid blockage due to human activities (\exempli engagement or obstruction) during navigation. More recently, some works employ LSTM-based models~\citep{alahi16-socialLstm}, GANs~\citep{gupta18-socialGan} or attention mechanisms~\citep{vemula18-attention} to condition trajectory prediction on intent estimates. This line of research has recently been observed in the neighboring field of autonomous driving where behavior prediction is often conditioned on goal estimates of vehicles or pedestrians~\citep{kris-2012, Yagi-2018,Zhang-2019, Liang-2019,rhinehart2019precog,Liang-2020,drogon}.

\subsection{Open Problems and Directions for Future Work}

Despite substantial and sustained interest in  planning framework design for social robot navigation, several fundamental problems remain.

\subsubsection{Modeling Interaction}

Much of the motivation for coupled prediction and planning stems from the inability of a robot to move through a dense crowd unless the robot accounts for human cooperation~\citep{trautmaniros}. However, the ``limit'' nature of the freezing robot problem---limit in the sense that it only applies in dense enough crowds---provides the impression that problems only arise after a density threshold is surpassed.  This is not necessarily true: even for 1 robot moving past 1 human, if the human is close (in velocity or position space) to the robot, then the agent density is high.  Thus, for a robot that ignores coupling, a nearby human can cause overly evasive maneuvers (or stopping without reason) that can make navigation unsafe.  This suggests that human-robot coupling (or, equivalently, coupled prediction and planning) is much more than just a ``limit'' phenomenon, but instead a core abstraction for social navigation architectures. Note that recent research in neighboring fields such as autonomous driving and human motion prediction is also dominated by approaches that strive to leverage models of coupled interaction~\citep{tang-2019,Roh2020Multimodal,Li19,chandra19,mavrogiannis2020implicit,li20,rhinehart2018deep}.

Unfortunately, coupled models---where each agent's movement is dependent on and influences every other agent's movement---are, in general, computationally intractable.  This complexity is easily visualized: consider $\nt$ agents, each with a discretized space of actions along 8 planar directions.   For a prediction horizon of $T$ timesteps, the system has $\calO(8^{\nt T})$ states. Even for simple cooperative collision avoidance objectives (such as described in~\citep{trautman-icaps-2020}), $10^6$ Monte Carlo joint samples consistently produce unsafe and inefficient solutions~\citep{trautman-icaps-2020}; this is hardly surprising: the prediction horizon is 3 seconds ($T=20$) and $\nt=7$, so the joint space is of size $8^{20\times7} = 8^{140}$. Importantly, this Monte Carlo scheme is not \emph{random}---samples come from probability distributions governing individual agent intent. What this suggests instead is that reasoning over the \emph{decoupled} space of agent intents will not recover a feasible \emph{joint} solution.  Although~\citet{trautman-icaps-2020} provided an efficient means to recover local optima by optimizing according to the joint objective, the structure of the global optimality landscape was left as an open question, precisely because general tools to analyze non-convex objectives do not exist.  Based on this result, we expect that joint objective functions for sampling based motion planning (\exempli PRM~\citep{prm}, RRT~\citep{rrt}, or RRT*~\citep{rrt*} based frameworks) will fail in much the same way, since there does not exist a method to draw samples from the joint objective---only from the individual agent models.  Even if sampling based motion planning increased performance by multiple orders of magnitude, this still would not be enough to overcome the complexity of this objective function.        

Broadly, we expect---and it has been proven---that coupling in existing planning formalisms---\exempli interactive POMDPs~\citep{ipomdp-complexity,intention-aware-crowd}, coupled dynamical systems~\citep{Warren2006} (in the best case, social navigation is related to n-body classical physical simulation, which requires supercomputers~\citep{coupled-dynamical-system-complexity}), game-theoretic approaches~\citep{nash-complexity,fridovich2020ilqr}, braid theoretic~\citep{mavrogiannis_ijrr}, or joint probabilistic inference~\citep{trautman-icaps-2020}---is inherently non-convex (whether some objective function might reduce the computational complexity is an open and important research question).  In general, for continuous-time Bayesian networks (similar to crowd navigation), complexity is NP-hard~\citep{cooper90,inference-ctbn}.   For reinforcement-learning approaches, \exempli \citet{chen17-sa-cadrl,chen19-crowdNav}, there is no reason to expect this complexity issue to disappear; rather, we expect that the complexity will be reflected in training burden. For learning-based approaches---\exempli imitation learning~\citep{social-nav-gan-end-to-end}---we would likewise expect data needs to reflect state space complexity.

Beyond the non-convexity of ``simple'' interacting agents (e.g. physical particles governed by gravitational equations), social navigation is also multi-objective (at the very least, we are typically concerned with balancing safety and efficiency).  The trade-off between safety and efficiency in coupled systems likely involves a Pareto front.  That is, not only is this problem non-convex, but there are likely many global optima with the same objective value.  This suggests that optimizing for safety and efficiency is an under-specification, and so additional constraints will be required.

This motivates elevating multi-objective non-convex optimization to the forefront of algorithmic social navigation research. We simply cannot provide basic guarantees that a coupled algorithm is even doing what we \emph{intend} it to do (apart from whether or not our objective or our data set appropriately reflects reality) until we can prove that a) we are acting according to a global optimum and b) if a Pareto front exists, the chosen optimum trades competing objective criteria in the desired manner.





\subsubsection{Behavioral Assumptions}\label{sec:behavioral-assumptions}

A significant design decision for any planning framework involves the assignment of a behavioral model for co-navigating humans. As discussed in Sec.~\ref{sec:movingobstacles}, early works in the field made the assumption that other agents move as dynamic obstacles with primitive or nonexistent models of intelligence. This turns out to be problematic in practice. Humans tend to attribute intentions to observed actions of others~\citep{Csibra07}. In fact, on many occasions, humans tend to expect cooperation in navigation~\citep{Wolfinger95}. A robot that does not abide by human expectations tends to confuse humans and distract human activity. On the other hand, pedestrian environments tend to feature a much richer selection of human activity: oftentimes, we encounter distracted pedestrians (\exempli looking at their phone), pedestrians in a rush, pedestrians with different perceptions of personal space, etc. 

Enabling a robot to robustly navigate in such rich environments would require the development of higher-level models of behavior. Much like humans, a competent robot must be able to adjust its behavior based on various types of human pedestrian behavior. 
Existing works often make the assumption of homogeneity, \idest all agents are assumed to be making navigation decisions by using the same exact planning mechanism. Further, several works still often make the assumption that others move \emph{linearly} by propagating their current velocities forward at every timestep \citep{chen17-sa-cadrl,Mavrogiannis18}. Moving forward in terms of algorithm design would inevitably require leveraging more high-fidelity, high-granularity models of high-level behavior, capturing richer aspects of human decision making.

\subsubsection{The role of context}
The majority of prediction and planning approaches do not explicitly model the context in which the robot is deployed. However, we are aware of the importance that the shape of a space (\exempli open space or narrow hallway), the timing (\exempli rush hour in a metro station or a museum) or the semantic map (\exempli at airports there are often queues next to the gates) plays on the type of crowd interactions. While different types of contexts such as environmental or task-specific contexts have been studied to design general intelligent agent behavior~\citep{schaefer2019integrating}, contexts has not been fully explored in the social robot navigation research. While there is work that incorporates environment geometry constraints to human behavior prediction~\citep{sadeghian19-sophie}, more emphasis should be placed to enable robust adaptation to different domains.

\section{Core Behavioral Challenges of Social Navigation\label{sec:behavioral}}




As Sec.~\ref{sec:behavioral-assumptions} suggests, pedestrian behavior modeling is crucial to successful navigation in rich environments. In order to reach the level of abstraction required for developing realistic behavioral engines, it is important to understand the behavioral implications of interaction in pedestrian environments. To this end, the body of work reviewed in this Section puts a heavier emphasis on predicting the behavioral implications of the robot and the pedestrians, instead of treating them as agents in a planning framework.

Extending beyond coupled prediction and navigation models, the human-robot interaction (HRI) community has worked towards analyzing key behaviors that express rich social signals in pedestrian environments. These social signals offer the planning algorithms hints such as which areas in the navigation space are considered socially inappropriate to trespass. They also offer additional insights on how pedestrians might react to the robot. 

We partition the works in three sections with increasing order of social signal richness and magnitudes. The first section is on proxemics, or personal spaces, which express signals on levels of discomfort between individuals. The second section is on intentions, which contain richer signals in pedestrian behaviors regarding timing, motions and predicted actions. The third section is on formations and social spaces, which consider signals inside a spatial arrangement of pedestrians or empty spaces among groups.

\subsection{Proxemics}

Proxemics describes a space around a pedestrian that, if intruded, might cause the pedestrian to feel discomfort. It is first noted by \citet{hall-1963,hall-1966} and signals the robots that they should not get too close to the pedestrians.  Early work focuses on simple models that consider each pedestrian individually. \citet{hall-1966} classifies personal space into concentric circular zones. Each zone represents a different level of affinity. Later, egg shape models were proposed by \citet{hayduk-1981} and \citet{kirby_thesis} from the observation that pedestrians value frontal space more than their personal space in other directions. The parameters of the egg shape models are then experimentally determined by \citet{barnaud-2014}. Later works believe that personal spaces are asymmetrical \citep{was-2006, gerin-lajoie-2008} and dynamic \citep{hayduk-1994}. \citet{Amaoka-2009} additionally considers gender and age when modeling personal spaces.

Regarding interaction with a robot, some works on proxemics focus on stationary subject approaching behavior. \citet{takayama-proxemics} studied the spatial signals around humans in both human-approach-robot and robot-approach-human scenarios. They analyzed which factors from the environmental context can affect the humans' proxemics behavior. \citet{Torta-2011} designed an empirical expression from user study experiments to model human's proxemics zones when approached by a robot. \citet{mead-probabilistic} used a probability based model to model social signals around stationary pedestrians when approached by a robot. These social signals are later used to analyze gesture and speech behavior \citep{Mead-2016}. 

In practice, proxemics signals are sometimes used as supplementary components to social navigation models. In a loose and simple form, they are manifested as potential fields around pedestrians. Social force based models \citep{Helbing95} define repulsive forces from other pedestrians. This can be viewed as one form of potential field based proxemics modeling. In \citet{svenstrup-human}, rapidly-exploring random trees are combined with a potential field \citep{potential-field,potential-field-limits} using parameters based on proxemics. \citet{clemson_nav} take a similar proxemics potential function based approach.

Other works treat proxemics signals as hard boundaries around pedestrians that should be respected. \citet{Lam-2011} defined various `sensitivity fields' with hard-boundaries that allow the robot to identify its social situation and execute the appropriate navigation strategy. \citet{Pacchierotti06b} defined a lateral passing distance in scenarios where a robot and a pedestrian walk head-to-head towards each other. \citet{Pandey-2010} used a two-level proxemics model to determine the reactive strategies for a robot to navigate around humans in which the robot should completely avoid these proxemics regions. In \citet{bera17-sociosense}, the authors used data derived parameters to define a dynamic rigid personal space and social space around pedestrians. 

Last but not least, some works model proxemics rules as a term inside a cost function to be optimized. In \citet{Chung-2010}, the authors optimize an overall cost function that consists of several weighted cost terms in order to generate trajectories for the robot to follow. One of the cost terms is defined by proxemics based rules. In \citet{Khambhaita-2020}, such an optimization is performed for both the robot and an interacting pedestrian reciprocally.

\subsection{Intentions}

Intentions often represent rich signals that offer hints to the future behavior of an agent. Literature from psychology and cognitive science~\citep{Csibra07,iseewhatyoumean,chrisbaker} has shown that humans tend to interpret observed actions as goal-directed. Therefore, by designing robot motion that is indicative of the robot's goal could enable an observer to infer the robot's future behavior. This has been shown to result in lower planning effort for a human agent in a navigation domain~\citep{Carton2016}, a finding indicating that legible robot motion may yield higher comfort for the co-navigating human. Note that other modalities such as body posture, eye gaze, arm/hand gestures and facial expressions could also provide rich intention signals. However, in this review, we focus on works that extract intentional signals from the motion of the robot.


Recently, there has been rich interest in the area of legible motion generation for human-robot interaction applications. \citet{dragan-2013} defined legibility as the property of motion that enables an observer to quickly and confidently predict the goal of an actor. \citet{Lichtenthaler-2012} showed that legible motions increase the perceived safety by pedestrians. \citet{kruse2012legible} designed a human-behavior imitating cost function to be optimized in order to produce legible motions for mobile robots. \citet{Szafir-2014} designed a set of expressive curvatures for an aerial robot to follow in order to signal its navigation goals. \citet{Mavrogiannis18} used the physical quantity of angular momentum as a heuristic that signals intent over a collision-avoidance strategy in multiagent environments.


In addition to future location hints, intentions offer more implicit, context-level signals that can further aid pedestrian's interpretation of robot motions. In a shopping mall setting, Satake et al. \citep{Satake-2009, Satake-2013} studied strategies for robots to successfully approach shoppers and ask them whether they need assistance. The authors found that the robots needed to successfully signal the pedestrians that they are being approached, or the pedestrians would miss the robot. \citet{Saerbeck-2010} used various combinations of different accelerations and path curvatures to invoke emotional interpretations from pedestrians. Similarly, \citet{knight-2014} produced expressive robot motions by drawing inspiration from Laban efforts. These expressive motions conveyed the robot's contextual state, such as being happy, or in a hurry.



\subsection{Formations and Social Spaces}

Spatial behavior of the pedestrians and their encompassed social spaces contain rich social signals. These signals guide a robot by informing the robot which social areas are inappropriate to trespass and by offering the robot additional insights to help with pedestrian motion tracking and predictions.

Most popular research on pedestrian spatial behavior is on grouping. Grouping is thought by some as a synchronization process where a few individuals attune their behavior to that of the people around them. Some \citep{Lorenz-2011, Iqbal-2016, Iqbal-2017} have investigated how robots can adjust their behavior to fit into human groups. From the perspective of navigation, researchers are more interested in groups rather than the grouping processes. Current works on groups can be divided into static and dynamic settings.

In static groups, people do not move and form interactions, such as conversation groups. Social space formed within static group formations should be respected by mobile robots. \citet{Vazquez-2015} and \citet{Vroon-2015} have studied how a robot may affect the spatial behavior of static groups by performing certain actions. A common tool used to analyse static groups is the F-formation \citet{Kendon-1990}, which was used by \citet{Kuzuoka-2010} to also analyse how a robot's action can affect 
static group spatial arrangements. Apart from studying a robot's effect on static group spatial arrangements, researchers are also interested in detecting these groups. Although sometimes not explicitly related to navigation, these static groups should be counted as impassable spaces and should be avoided during navigation. Again, some work has used F-formations to detect these groups \citep{Vazquez-2015-det, Hung-2011, Setti-2015}. Recent work has also used deep learning based models to detect static groups from images and videos \citep{Alameda-Pineda-2015, Choi-2014}.

Dynamic groups are groups implicitly formed by moving pedestrians. Similar to static groups, social spaces are present within dynamic groups and are best not to be interrupted. Similar to static groups, \citet{Fiore-2015} and \citet{Shiomi-2007} have studied how robots can affect dynamic group spatial behavior, as commonly seen in museum tour guide robot settings. \citet{Garrell-2012} also studied appropriate robot spatial behavior when accompanying pedestrians. Apart from spatial behavior, most of the focus when studying dynamic groups is on detection or tracking. Detection and tracking of groups are examined from both exo-centric and ego-centric perspectives. 

From exo-centric perspective, common approaches include probabilistic progression based methods \citep{Bazzani-2012, Chang-2011-group, Gennari-2004, Pellegrini-2010-group, Zeynep-2013, Zanotto-2012}, graph-based approaches \citep{chamveha-2013, Khan-2015, Feng-2015, Noceti-2014}, clustering methods \citet{Ge-2012-group, Solera-2016}, and social force model based approaches \citep{Mazzon-2013, Sochman-2011, Leal-Taixe-2011}. From ego-centric perspectives, grouping becomes more difficult because sensor errors and pedestrian occlusions are present. Common models use Multiple Hypothesis Tracking models \citep{Mucientes-2006, luber-2013, lau-2010}, which focus on data association between group selection hypothesis on two consecutive frames, and clustering based approaches \citep{Chatterjee1, taylor-2020-group} on features abstracted from sensor inputs.

Apart from generic group analysis, some researchers focus on specific types of grouping behavior or actions, such as following behavior. Researchers studied how a robot should follow pedestrians \citep{Gockley-2007, Granata-2012, Jung-2012, Zender-2007} and conversely, how a robot should guide pedestrians \citep{nanavati-dyad, Feil-Seifer-2011, Garrell-2010, Pandey-2009}. Others focus on how a robot should stay in group formations \citep{Althaus-2004, Cuntoor-2012, Martinez-Garcia-2005}. Lastly, some researchers focus on predicting group splits and merges \citep{wang-split-merge}.

Researchers have also focused on meta-level formations in high density crowds. \citet{Henry-2010} looked at lane formations in narrow corridors when crowded pedestrians walk in opposing directions. \citet{van-den-Berg-2008} observed such behavior in simulation from their navigation model. \citet{Cheriyadat-2008} aimed to detect dominant pedestrian flows in surveillance videos in dense crowds.



\subsection{Open Problems and Directions for Future Work}

\subsubsection{Better behavior understanding}

Despite the large body of work on pedestrian behavior, the biggest behavioral challenge of social navigation is still on \emph{understanding} pedestrian behavior. We attribute this difficulty primarily to the shortage of principled knowledge of pedestrian behavior. At the current stage, research in pedestrian behavior understanding takes a trial-and-error based approach, where researchers design a model for a specific aspect of pedestrian behavior and often show in a small-scale lab study that the model shows promise. While the lab studies and validation methods used to establish scientific significance of these models are sufficient, lab studies are often heavily controlled and lack authenticity when compared to real world scenarios. Sec.~\ref{sec:lab-studies} provides more discussion on lab studies in social navigation model evaluations. To turn these models into principled knowledge, they need to be tested extensively in more generalized, real-world environments.

Limited principled knowledge is also reflected in recent research trends, where researchers tend to use deep learning to implicitly model pedestrian behavior and directly output behavioral outcomes such as grouping, trajectory predictions, or navigation actions. While these approaches often produce impressive results, they do not produce explicit knowledge of pedestrian behavior. This knowledge is needed for a robot to imitate pedestrian behavior, because it is equally important to understand pedestrian behavior and for a robot to act like a pedestrian in order to be perceived as being safe and trustful. 

\subsubsection{Behavior importance evaluation}
Pedestrian behavior analysis is a conglomeration of the various behavioral aspects that potentially benefit social navigation. Researchers often focus on one of these aspects to explore. However, it is often unclear how much the inclusion of such particular behavioral aspects benefit social navigation. For many of the works mentioned above, such as group detection and trajectory predictions, the authors have yet to integrate with social navigation. Future works are recommended to integrate social navigation planners with these behavioral aspects in order to analyze their practical effectiveness. 









\section{Core Evaluation Challenges of Social Navigation}\label{sec:evalchal}

Social navigation is a sequential decision making problem with multiple objectives as defined in Section~\ref{sec:problem}. In the navigation problem without the social interaction part as defined in Definition~\ref{eq:nav}, the primary objective of an agent is to reach their destination. Considering this objective only, an agent's navigation performance can be evaluated based on how successful in arriving at a destination, \exempli using the physical distance between an agent's position and a destination. Among the solutions that have reached a destination, it is generally subjective to rank alternatives according to multiple preferences or soft constraints.  Social compliance, for example, is a soft constraint that is comprised of many aspects including safety, comfort, naturalness, and other societal preferences. 
The quality of social navigation performance can thus be evaluated differently depending on where the focus is among those potentially competing objectives. 

In this section, we first review the metrics used to measure various facets of the social navigation performance. Next, we describe three evaluation methodologies: datasets, simulations, and real-world experiments. Finally, we discuss remaining challenges on how to measure success in social navigation.  

\begin{table*}
\centering
\footnotesize\begin{tabular}{| p{4cm} | p{7cm} |}
\hline
Name & Description \\ [0.5ex] 
\hline
\hline
Arrival rate& Percentage of agents which successfully reach the goal \\ \hline
Path length& The distance traveled by an agent \\ \hline
Collision rate& Percentage of cases with at least 1 collision \\ \hline
Failure rate& Percentage of agents that fail to reach destinations \\ \hline
Average time to goal& Average time to reach goals \\ \hline
Social score & Success rate and comfort \\ \hline
Average acceleration & Acceleration averaged by time \\ \hline
Average energy & Integral  of squared  velocity  of an agent, averaged by timestep \\ \hline
Path irregularity & total amount of  rotation beyond straight line path \\ \hline
Path efficiency & Ratio between straight line path and actual path \\ \hline
Time spent per unit path length& Average time spent \\ \hline
Topological complexity & Amount of entanglement among agents' trajectories \\ \hline
Speed efficiency& Ratio of nominal and actual speed \\ \hline
 \end{tabular}
 \caption{Evaluation metrics for social navigation datasets.}
 \label{tbl:metrics}
 \end{table*}
 
\subsection{Metrics}\label{sec:metrics}

The metrics discussed in this section are for evaluating the performance of a single agent. 
We note that some of these metrics in aggregation can also be used to evaluate a population performance. For instance, such statistics collected from the human pedestrian datasets can be used as the references for measuring how closely an algorithmic population, \idest an algorithm in self play, resembles human pedestrians in their navigation behavior. 

The metrics described in subsections are summarized in Table~\ref{tbl:metrics}. The statistics of five commonly used pedestrian trajectory datasets according to such metrics are shown in~Table~\ref{tbl:metrics-evaluation}.

\subsubsection{Navigation success rate: }\label{sec:succesrate}
The arrival rate measures how often an agent reaches its a goal that is a primary metric in general robot navigation~\citep{oh2016-iser} and has also been used in social navigation~\citep{tsai20}. In general, some assumptions are added to measure the arrival rate. For instance, a time constraint can be added to enforce an agent to reach its destination within a certain deadline. In the case of datasets, because the true destinations of pedestrians in a pedestrian dataset are generally unknown, people's destinations need to be assumed. For instance, the location from which an agent exits the scene can be considered as the agent's final destination for the purpose of computing the arrival rate, where the locations can be represented in either a discrete~\citep{vemula17-igp} or continuous space~\citep{tsai20}. 

The arrival rate depends not only on an agent's navigation algorithm but also on its environmental context such as the crowd density. Combined with the average speed of agents, the arrival rate can be used to measure the level of congestion of an environment. 

\subsubsection{Path efficiency and optimality:} \label{sec:path-efficieicy}

Whereas the success (or arrival) rate evaluates the performance based on the final outcome, the path-efficiency metrics measure the quality of the trajectories. For instance, the length of a path~\citep{trautmaniros} and
the time needed for an agent to travel $6$ meters~\citep{trautman-ijrr-2015} have been used to measure the efficiency given a path. 

\citet{Mavrogiannis19} proposed a set of metrics for measuring various aspects of trajectory quality. The average acceleration of an agent and the average energy (integral of squared velocity over a trajectory segment), measure subtle changes in pedestrian motions. The path irregularity metric (amount of unnecessary rotation), modified from the work of \citet{Guzzi}, and the path efficiency metric (ratio between Euclidean distance to goal and the actual traveled distance) measure the level of disturbance in pedestrian behavior.  
Finally, the time spent per unit length, the reciprocal of average speed, and the speed efficiency measure different aspects of efficiency of a trajectory. 

The ratio of nominal and actual speed has also been proposed in~\citep{sim2real-crowds}.

\subsubsection{Safety / Collision Avoidance:}

The safety performance can be measured in terms of the number of constraint violations, for instance, the number of collisions~\citep{fraichard-three-rules}. In addition to hard constraint violations, continuous metrics have been proposed to define 
a robot's closest distance to a human, known as the safety margin~\citep{trautmaniros} or the minimum distance~\citep{Mavrogiannis19}.

\subsubsection{Behavioral naturalness:}\label{sec:naturalness}

Naturalness is not directly connected to a planning performance, but the concept represents a fitness score for a robot to blend in a human environment.
    To evaluate how well an agent's behavior resembles a human's, a dataset of recorded human pedestrians can be used. 
    It is a common method to evaluate the naturalness of a robot navigation in terms of the distance between an agent's plans and the actual trajectories taken in the recordings. 
    Two distance metrics that are prominent in existing works are 
    the average displacement error and the final displacement error. Naturally, these metrics are also predominant in the pedestrian prediction literature~\citep{rudenko2019-predSurvey}.
    
    Average Displacement Error (ADE) introduced in ~\citep{walkalone} is the average of Euclidean differences between temporally aligned points in the two paths given by a navigation algorithm and the ground truth path from the dataset, respectively. ADE has been used in~\citep{yao19-group} and~\citep{hamandi19-deepmotion}. 
    
    Final Displacement Error (FDE) proposed in ~\citep{alahi16-socialLstm} measures the displacement error only at the final time step. In this regard, it is related to the arrival rate, although the notion of a goal is only implied here. 
    
    The prediction accuracy~\citep{bera17-sociosense} is the percentage of successful predictions in all attempted predictions. A success in this context is typically defined based on a threshold value and another similarity metric such as the average or maximum displacement error, such that a prediction is regarded as successful if its similarity to the ground truth value is higher than the threshold. 
    
    Dynamic Time Warping (DTW) used in~\citep{hamandi19-deepmotion} finds the optimal way to warp prediction to a ground truth path, and considers the cost of performing such a warping as the difference between prediction and truth. In the similar vein, the Hausdorff distance or Fr\'echet distance can also be considered.  
    
    \citet{tsai20} proposed the comfort rate following the theory of social force~\citep{Helbing95}. The original social score definition combines the safety score with the navigation score, \exempli whether an agent has reached the destination. The comfort score only takes the social compliance term into consideration based on the minimum required social distancing constraint. 
    
    
    \citet{Mavrogiannis18,Mavrogiannis19} employed the topological complexity index \citep{Dynnikov2007} as a measure that quantifies the intensity of \emph{mixing} observed in the motion of multiple navigating agents. The higher the complexity, the more agents directly confront each other as they navigate towards their destinations. Topological complexity was shown to be correlated with a notion of \emph{Legibility} \citep{DraganAuR14} in multiagent navigation scenarios.

\subsection{Evaluation Methodologies}

In this subsection, we discuss the methodologies used in literature and their pros and cons in evaluating various facets of social navigation. 

\subsubsection{Datasets} 

Using a dataset, evaluation is limited to measuring the difference between predictions and the recorded behaviors using a metric such as displacement errors. There are several issues here: 1) recorded behaviors may have been conditioned to a certain context; 2) alternative behaviors that are different from the recording but are equally good can be unfairly penalized; 3) the majority of existing datasets do not have the notion of destinations for the agents in a scene, making it less suitable for a navigation task; and 4) recorded behaviors rarely include hard constraint violation such as collision and these rare events need to be engineered/trained based on domain knowledge. 

There are two datasets widely used in both social robot navigation and human motion prediction literature: ETH~\citep{walkalone} and UCY~\citep{Lerner1}. The ETH dataset contains trajectory data from two scenes, namely \emph{ETH} and \emph{Hotel}, while the UCY dataset contains \emph{Univ}, \emph{Zara1}, and \emph{Zara2}. \emph{ETH} and \emph{Hotel} have more linear trajectories and lower pedestrian density when compared to others. \emph{Zara1} and \emph{Zara2} have relatively more complex trajectories and higher pedestrian density. \emph{Univ} has the most non-linear trajectories and the highest density out of these five scenes. 


\begin{table*}
\centering
\resizebox{\textwidth}{!}{
\begin{tabular}{| p{5cm} || p{2cm} | p{2cm} | p{2cm} | p{2cm} | p{2cm} |}
\hline
Name &              ETH & HOTEL & UNIV & ZARA1 & ZARA2 \\ \hline \hline
Avg. time to goal ($sec$) & 0.253  &  0.770 & 2.404& 0.628  & 1.089 \\ \hline
Avg. acceleration ($m^2/sec$)  &   0.092   &   0.044& 0.027  &    0.019& 0.023 \\ \hline
Avg. energy ($m^2/sec^2$) &     6.296&  1.765   &0.747&   1.445&  1.473 \\ \hline
Path irregularity ($degree/m$) & 0.914 & 3.973 & 3.708 & 2.202 & 4.702\\ \hline
Path efficiency ($\%$) & 0.983 & 0.933 &0.912 &  0.981 & 0.969 \\ \hline
No. of agents           &   360     & 389   & 967   & 148 & 204 \\ \hline
Avg. Agent Duration ($sec$)    &   5.63    & 5.80  & 15.04 & 13.53 & 18.66 \\ \hline
Avg. Pedestrian Density ($./m^2$)  & 0.313   & 0.312 & 0.398 & 0.354 & 0.384 \\ \hline
Max. Pedestrian Density ($./m^2$)  & 0.583   & 0.750 & 0.750 & 0.625 & 0.75 \\ \hline
Min. Pedestrian Density ($./m^2$)  & 0.250   & 0.250 & 0.250 & 0.250 & 0.25 \\ \hline 
 \end{tabular}
 }
 \caption{Dataset comparison according to the metrics in~Table \ref{tbl:metrics}.}
 \label{tbl:metrics-evaluation}
 \end{table*}
 
    Using pedestrian datasets for evaluation is favored due to two reasons. First, although the recorded behaviors are not fully representative of human pedestrians, the datasets still contain the trajectories of humans navigating in a real environment. Second, it is practically much more convenient to evaluate on a dataset when compared to testing in a real human environment.
    The dataset-based evaluation, however, presents several limitations. 
    These metrics are only applicable when the evaluation is done by comparing navigation algorithm outputs to dataset recordings. This way of evaluating navigation algorithms carries strong prior assumption about the optimality of human behavior recorded in dataset, while in reality paths chosen by human pedestrians might not be optimal or there are likely more than one optimal paths for the same scene. Forcing navigation algorithms to give the exact same path as recorded ground truth may not be wise. 


\subsubsection{Simulation}


Real-life experiments are usually expensive in both time and resources. 
Crowd simulators can be easily customized to support new tasks and environments in a way that would not be otherwise possible in the real world. 
They are essential tools to evaluate social navigation algorithms and train data-driven or reinforcement learning models~\cite{chen19-crowdNav, chen17-sa-cadrl}.
As \citet{sim2real-crowds} point out, existing crowd simulators rely on unrealistic assumptions that make them far from being a high-fidelity representation of the real world.
They assume that all agents obey the same law (\emph{homogeneity}), and are aware of the state of all other neighboring agents (\emph{omniscience}). Additionally, in simulations, virtual agents are allowed to collide as long as these collisions do not impact visual realism (\emph{safety}). 
Nevertheless, once these assumptions are relaxed, we run into safety issues. It is impossible to guarantee that a policy trained in such simulated environments and then transferred into the real world will not cause the robot to collide with other human agents. 


\paragraph{Models}

The literature proposes several techniques to model the crowd's dynamics, where pedestrian simulation engines are drawn from work in crowd analysis dynamics~\citep{helbing1} or computer graphics~\citep{vandenberg,upl,vbn}.
Existing works mainly focus on simulating crowd behaviors in a critical context, such as the circular aggregation or flow of pedestrians around a narrow passage and lanes' formation between two or more pedestrians' flows~\citep{helbing2000simulating, Helbing95, paris2007pedestrian, ondvrej2010synthetic, guy2012least} or on designing more artificial scenarios to test interactions between agents~\citep{van2008reciprocal,ondvrej2010synthetic, guy2009clearpath}.

Although it is not common to observe these kinds of scenarios in real life, they provide the desired situations to evaluate collision avoidance algorithms in progressively changing density conditions.
A crowd simulator must achieve sufficient generality and reach high complexity and randomness to avoid overfitting and guarantee the policy's convergence during training.
There are three main approaches to model the crowd dynamics, \idest  microscopic, macroscopic, and mesoscopic models. See Table~\ref{tab:taxonomy-models} for a taxonomy of widely employed methods.

The microscopic model attempts to study crowd behavior by modeling each individual's action through force-based or velocity-based models. The Social Force Model (SFM)~\citep{helbing2001} is one of the most prevalent models in this category. It is a force-based method, simulating crowd dynamics through interactions between agents and obstacles, where both socio-psychological and physical forces influence the individual decision-making process.
It was first introduced to study the crowd dynamics in a situation of escape panic.
Other works build upon the SFM algorithms by introducing a mobile grid to make each agent adjust its preferred velocity~\citep{saboia2012crowd}, resulting in more smooth trajectories, or adding evasive forces to predict future collisions~\citep{karamouzas2009predictive} and interactions, and reduce particle oscillations~\citep{ji2016vpbs}.
Under the velocity-based modeling approaches, we can find techniques such as Velocity Obstacle (VO)~\citep{fiorini1998motion}, Reciprocal Velocity Obstacle (RVO)~\citep{van2008reciprocal}, Optimal Reciprocal Collision Avoidance (ORCA)~\citep{vandenberg}, and Hybrid Reciprocal Velocity Obstacle (HRVO)~\citep{snape2010smooth}.
These algorithms generate collision-free trajectories by considering combining local interactions and the neighbor's information into the decision-making process. 

The macroscopic approach models the crowd dynamics at a large-scale. The crowd is represented as a continuous entity, where potential fields and fluid dynamics models govern the pedestrian motion.  This modeling approach works very well to describe high-density situations, where it is possible to recognize the groups' appropriate behavior.
However, it does not consider individual-level interactions between agents and the environment.
Under this category, there are mainly three ways to design pedestrian flow dynamics, \idest  continuum theory, aggregate dynamics, and potential field.
The continuum model~\citep{hughes2002continuum, jiang2010continuum} describes the pedestrian flow dynamics through the use of the continuum theory. The crowd is a flow characterized by density and speed. This way of modeling crowd behavior, however, is more designed for large groups with common goals. The aggregate dynamics model~\citep{narain2009aggregate} uses a fluid dynamical model to represent the crowd flow and discrete agents or continuous systems to describe the crowd density.
The potential-field method~\citep{patil2010directing} aims to model one or more agent actions by integrating guidance fields designed by a user to generate smooth, collision-free navigation fields.

Recent approaches investigate mesoscopic models to design group simulations rather than single or large-scale models. This method includes dynamic group behaviors~\citep{dion2000group, lewin1945research, karamouzas2011simulating} (\idest the social relationships among individuals), which uses dynamic analysis to obtain general rules of group phenomena, interactive group formation~\citep{ulicny2004crowdbrush, yersin2009crowd}, and social-psychological crowds~\citep{guy2011simulating, sakuma2005psychological, wiggins1996five, eysenck1950dimensions}, used for emergency evacuation and parades. The latter model considers personality traits and emotion contagion theories and how they affect human behaviors into the modeling framework.

\paragraph{Software}

This section reviews some relevant crowd simulator software available as research platforms for crowd behavioral studies. 


\emph{PedSim}~\citep{pedsim}, \emph{OpenSteer}~\citep{reynolds1999steering}, \emph{Menge}~\citep{menge}, \emph{Continuum}~\citep{treuille2006continuum}, both written in C++, are the most common tools used to simulate crowd scenarios.
\emph{PedSim}~\citep{pedsim} is a C++ library based on the social-force-based steering model. One of this method's main drawbacks is that it has no global planning ability neither high-level behaviors and goals.
\emph{OpenSteer}~\citep{reynolds1999steering} is a C++ application used to explore steering behaviors for vehicles but not specifically designed to simulate human pedestrians (\idest the human agent derives from a ``vehicle'' class).
Both these tools, however, have limited capabilities. They do not account for different human behaviors and have no global planning capability. \emph{Menge}~\citep{menge} is a C++ modular framework with the ability to use run-time plug-ins to change the pedestrian model, global navigation algorithms, and the way we can visualize the simulation. 
Fluid dynamical models have inspired other approaches to simulate crowds.
The Continuum Crowd~\citep{treuille2006continuum} is a tool based on continuum dynamics and dynamic potential field.  This software is based on the continuum model and represents pedestrians as a continuous density field where partial differential equations describe their dynamics.
As a drawback, none of this software considers the robot as part of the crowd scenario.

\emph{PedSim-Ros} builds upon~\cite{pedsimros} to include a robot agent in the simulation environment. 
\emph{Webots},  \emph{ROS} (Robot Operating System)~\citep{quigley2009ros}, \emph{Gazebo}~\citep{koenig2004design}, and \emph{Stage} \citep{gerkey2003player}  are other frameworks used to represent virtual environments, where each agent is defined by its geometry and dynamics and navigation strategy.  These tools, however, lack in both graphical and physical realism.
\emph{MengeROS}~\citep{aroor2018mengeros}, is an open-source simulator simulate a variety of human crowd behavior(force-based, velocity-based) and by integrating the robot in the scene.

\emph{CrowdSim}~\citep{cassol2016crowdsim}  is a rule-based crowd simulation software developed to simulate pedestrian behaviors during evacuations, along with human comfort and safety. It is an add-on for \emph{Blender}~\cite{blender}, and it is used primarily for visual effects. This software offers the possibility to generate random test data with high variability and controlled complexity.
\emph{UCrowds}~\citep{ucrowds} is a simulator available as 3D Plugin for \emph{Unity}~\citep{unity}. Its primary objective is to simulate the movements of large masses of people under a variety of circumstances. It offers faster and more accurate simulations and the possibility of producing dynamic adjustments, making it very appealing for crowd behavioral studies. 

\emph{Golaem Crowd}~\citep{Golaem}, \emph{Massive}~\citep{massive}, and \emph{Miarmy}~\citep{miarmy} are photo-realistic simulators available as a plug-in for \emph{AutoDesk Maya}~\citep{maya}. They are explicitly used for generating special effects and visualization, but they are not suitable for crowd behavioral studies.

Lastly, an experimental platform namely \emph{SEAN-EP}~\citep{tsoi2020sean} has been introduced as a high visual fidelity and open-source system social navigation simulation platform, built on top of Unity, that evaluates and gathers feedback on navigation algorithms.

\begin{table}
  \centering
  \scalebox{1}{
  \begin{tabular}{|c|c|c|c|}
    \hline
    \textbf{Model} & \textbf{Microscopic} & \textbf{Macroscopic} & \textbf{Mesoscopic} \\
    \hline
    \hline
    \citet{Helbing95}  & \checkmark  &   & \\ \hline   
    \citet{saboia2012crowd}  & \checkmark  &   & \\ \hline
    \citet{karamouzas09}  & \checkmark  &   & \\ \hline
    \citet{fiorini1998motion}  & \checkmark  &   & \\ \hline
    \citet{van-den-Berg-2008}  & \checkmark  &   & \\ \hline
    \citet{snape-hrvo}  & \checkmark  &   & \\ \hline
    \citet{hughes2002continuum}  & \checkmark  &   & \\ \hline
    \citet{Treuille2006} &   &\checkmark   & \\ \hline
    \citet{jiang2010continuum} &   &\checkmark   & \\ \hline
        \citet{narain2009aggregate} &   &\checkmark   & \\ \hline
        \citet{patil2010directing} &   &\checkmark   & \\ \hline
        \citet{dion2000group} &   &   & \checkmark \\ \hline
        \citet{lewin1945research}  &   &   & \checkmark \\ \hline
        \citet{karamouzas09}  &   &   & \checkmark \\ \hline
                \citet{ulicny2004crowdbrush} &   &   & \checkmark \\ \hline
                \citet{yersin2009crowd} &   &   & \checkmark \\ \hline
                \citet{guy2011simulating} &   &   & \checkmark \\ \hline
                \citet{sakuma2005psychological} &   &   & \checkmark \\ \hline
                \citet{wiggins1996five} &   &   & \checkmark \\ \hline
                \citet{eysenck1950dimensions} &   &   & \checkmark \\ \hline
  \end{tabular}}
  \caption{A Taxonomy of Crowd Models.\label{tab:taxonomy-models}}
\end{table}

\subsubsection{Experimental Evaluation}\label{sec:expeval}

A crucial part of evaluation in social robot navigation research involves testing a navigation framework in close proximity with humans. In this section, we classify existing work into categories depending on the experimental methodology followed for validation. Broadly, we observe that researchers have followed three main approaches: a) experimental demonstrations presenting proof-of-concept prototypes; b) lab studies under controlled settings; and c) field studies in public environments. Table~\ref{tbl:littaxonomy} presents a taxonomy of existing works based on the validation methodology followed.

\paragraph{Experimental Demonstrations}

An experimental demonstration in the presence of humans often represents the first real-world step following a simulated case study for social robot navigation research. This type of evaluation often features limited control over experimental variables and limited reporting of qualitative or quantitative performance measures. For example, \citet{Bennewitz05} report a series of ten experiments involving robot navigation in a hallway of limited crowd density. \citet{Sisbot07TRO} descrinbe a series of navigation interactions between a robot running their proposed navigation algorithm and a human in a lab environment. \citet{park2012robot} test their control framework on a robotic wheelchair inside a corridor of an academic building and report a set of successful collision-avoidance encounters. \citet{kuderer-ijrr-2016} also deploy their model on a robotic wheelchair documenting a set of experiments at a narrow hallway under controlled settings. \citet{chen17-sa-cadrl} document an experimental demo, involving a robot navigation experiment in a crowded area of an academic building.

\paragraph{Lab Studies}\label{sec:lab-studies}

Lab studies provide the advantage of the ability to control experimental variables that play a significant role during execution. For this reason, lab studies represent an important step for design and evaluation of social robot navigation research. \citet{Pacchierotti06b} test their control framework with a study involving an autonomous robot navigating next to human subjects at a corridor under controlled settings. They present their findings from the interactions of 10 participants with a robot exhibiting different navigation strategies corresponding to different passing distance. \citet{kirby09} and \citet{kirby_thesis} present a user study involving 27 human subjects navigating alongside a robot in an academic hallway. \citet{kruse2012legible} document a series of interactions between a mobile robot and a human subject in a lab study involving 10 participants. \citet{truong} document a series of interactions between a robot running their planning algorithm and human participants in a lab environment. Our past work \citep{Mavrogiannis19} featured an experimental validation of our planning framework \citep{Mavrogiannis18} in a lab experiment involving interactions between a navigating robot and three human subjects at a time under dense navigation settings, yielding a total sample of 105 participants. \citet{Lo19} evaluate a series of robot collision-avoidance strategies on a self-balancing mobile robot in a lab study with 98 human subjects. 
\citet{tsai20} evaluated their approach on a physical robot in a controlled outdoor environment where they tested robot behavior against various settings of pedestrians. They used two types of  human evaluations based on the feedback from the evaluators observing the scene from a third-person view and the participants (pedestrians). 

\paragraph{Field Studies}
Another common approach involves deploying robots in the wild in public environments. \citet{rhino-robot} deployed the RHINO guide robot in the Museum of Bonn, in Germany in 1997 and documented thousands of in-person and virtual interactions between the robot and visitors over 47 hours or runtime spanning 6 days. \citet{thrun_ijrr2000} deployed MINERVA, the second generation of robot tour guides at a Smithsonian Museum in Washington DC, USA, in 1998 and also documented thousands of interactions with visitors for two weeks. Both studies documented statistics related to performance, collision avoidance and visitors' impressions of the robot.

\citet{fokamuseum} report logs and performance aspects upon running their robot for 70 hours in an indoor academic building. \citet{Shiomi2014} test their navigation framework through a 4-hour field study in a shopping mall. \citet{trautman-ijrr-2015} test their navigation framework on a robot in a field study comprising 488 robot runs in a crowded cafeteria. \citet{Kato15} test their approach on a humanlike robot employee in a crowded mall and record interactions with 130 people. \citet{Kim2016} evaluate their framework on a robotic wheelchair in a crowded hallway over 10 field runs.

\begin{table}
\centering
\scalebox{1}{
\begin{tabular}{| c | c | c | c |}
\hline
Study & Demo & Lab Study & Field Study\\ [0.5ex]
\hline
\hline
\citet{rhino-robot} &  &  & \checkmark\\
\hline
\citet{thrun_ijrr2000} &  &  & \checkmark\\
\hline
\citet{Bennewitz05} &  &  & \checkmark\\
 \hline
\citet{Pacchierotti06b} &  & \checkmark & \\
 \hline
\citet{Sisbot07TRO} & \checkmark &  & \\
\hline
\citet{ziebart2009} & & \checkmark &  \\
\hline
\citet{fokamuseum} &  &  & \checkmark\\
\hline
\citet{kirby_thesis} &  & \checkmark & \\
\hline
\citet{park2012robot} & \checkmark &  & \\
\hline
\citet{kruse2012legible} &  \checkmark &  &  \\
\hline
\citet{intention-aware-momdp} & \checkmark &  & \\
\hline
\citet{learning-social-robot} & & \checkmark &  \\
\hline
\citet{Shiomi2014} & \checkmark &  & \\
\hline
\citet{Kato15} &  &  & \checkmark\\
\hline
\citet{Kim2016} &  &  & \checkmark\\
\hline
\citet{trautman-ijrr-2015} &  &  & \checkmark\\
\hline
\citet{kuderer-ijrr-2016} & \checkmark &  & \\
\hline
\citet{chen17-sa-cadrl} & \checkmark &  & \\
\hline
\citet{truong} &  & \checkmark & \\
\hline
\citet{Mavrogiannis19} & & \checkmark & \\
\hline
\citet{Lo19} & & \checkmark & \\
\hline
\citet{tsai20} & & \checkmark & \\
\hline
\end{tabular}
}
\caption{A taxonomy of studies in algorithmic social robot navigation.}
\label{tbl:littaxonomy}
\end{table}

\subsection{Open Problems and Directions for Future Work}

In this section, we highlight a series of problems related to existing practices on the evaluation of social robot navigation frameworks and offer suggestions for improvement.

\subsubsection{Limitations of Existing Simulation Practices}\label{sec:experiments}

Simulation is a valuable tool for prototyping and testing social robot navigation frameworks in controlled settings. However, simulating crowd navigation is a challenging task on its own as it requires adopting assumptions that inevitably abstract away too much out of the rich interactions that naturally arise in real-world situations. Therefore, it is crucial to carefully construct an experimental setup that can provide meaningful insights. This requires determining the models governing the behavior of simulated humans, the specific scenarios in which we evaluate an algorithm, the metrics with respect to which we evaluate them, and more. 

In the absence of a uniformly agreed upon evaluation standard, existing evaluation practices found in the literature often involve strong and unrealistic assumptions that prohibit the extraction of meaningful insights. For instance, the majority of recent literature~\citep{chen19-crowdNav, chen2019relational, everett18-lstm, lisocially} relies on ORCA \citep{van2011rvo2} as the main behavioral engine governing simulated humans for both training and testing of learning algorithms. While this choice is motivated, it comes with serious shortcomings. We uncover these shortcomings with a simulation study.

\begin{figure}[h!]
     \centering
     \begin{subfigure}{.485\linewidth}
     \centering
         \includegraphics[width=\linewidth]{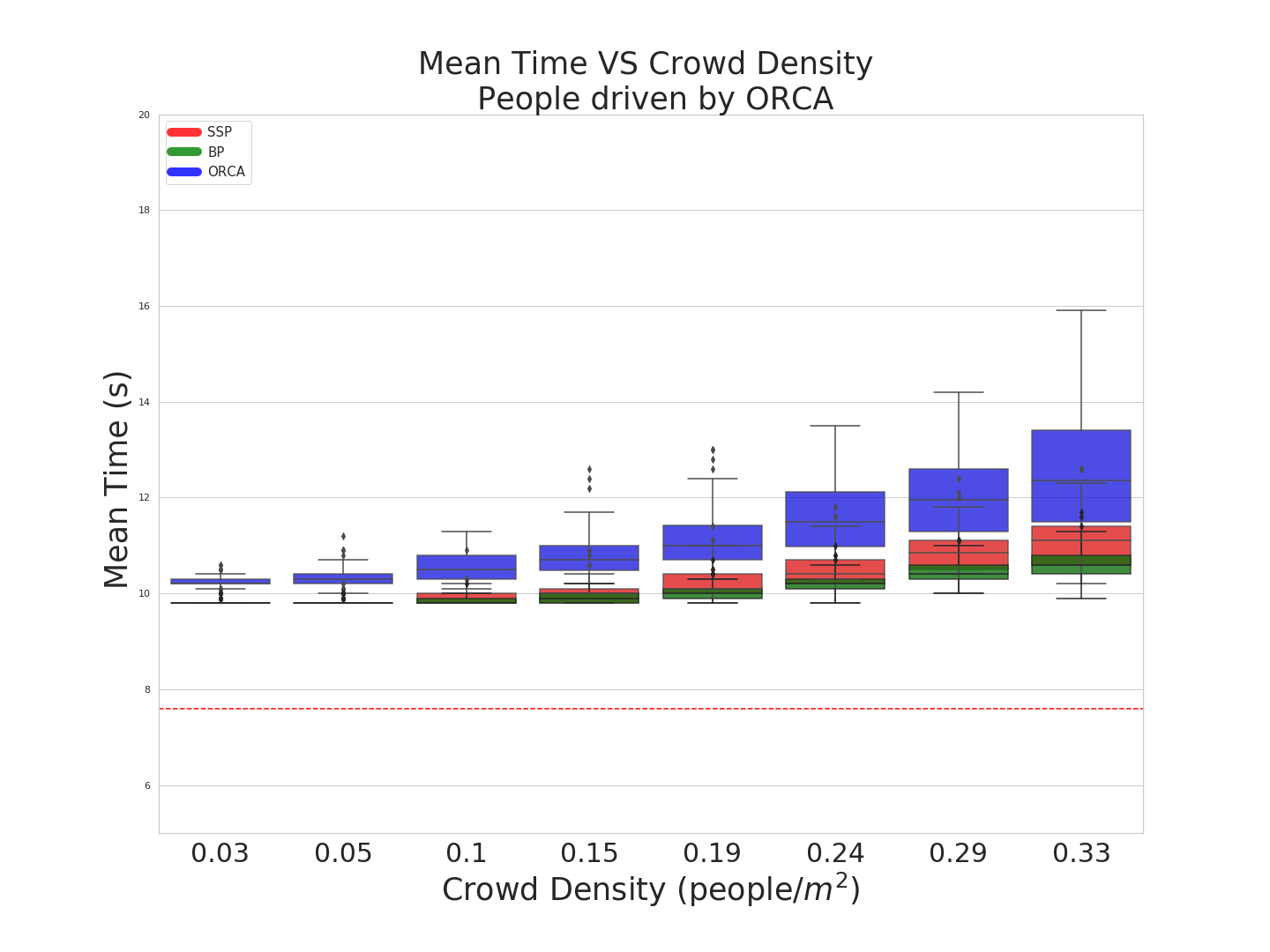}
         \caption{Average time to reach the goal on ORCA. The SSP or BP significantly outperforms an ORCA-driven robot in an ORCA environment. ORCA algorithm chooses velocity over paths, and so lacks a mechanism to represent the general navigation task; meanwhile, a naive planner such as SSP preserves this geometry information for path planning, achieving greater performance.}
         \label{fig:orca}
     \end{subfigure}
      \hskip 2ex
     \begin{subfigure}{.485\linewidth}
              \centering
         \includegraphics[width=\linewidth]{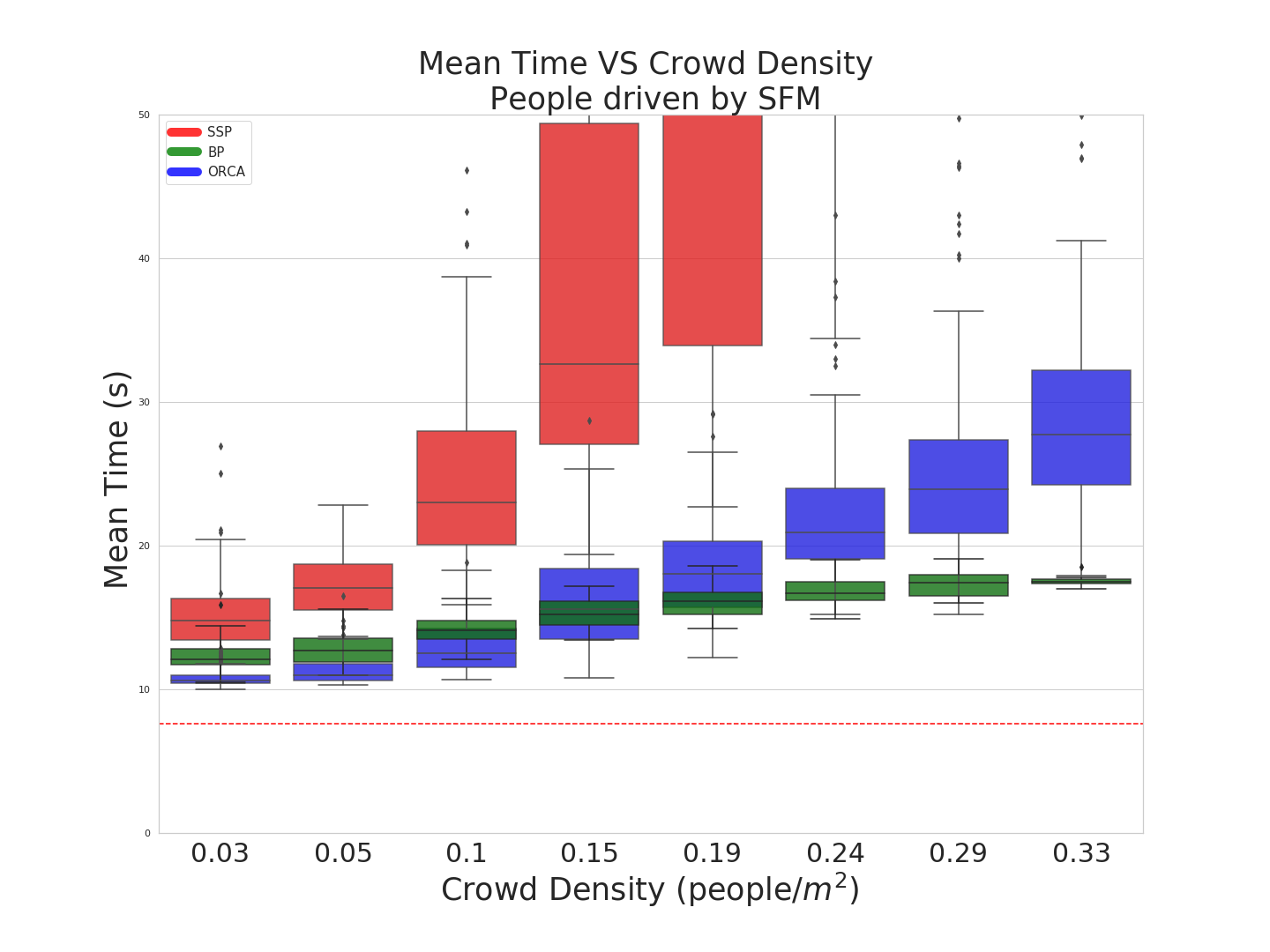}
         \caption{Average time to reach the goal on SFM simulator. In this scenario, the SSP underperforms compared to the other methods. It fails to reach the goal when crowd density over $0.2 people/m^2$  while the BP presents more than $70 \%$ of collision cases. This sort of behavior is the one that is more likely to appear in a real-world setting.\\\:}
         \label{fig:SFM}
    \end{subfigure}
    \caption{Average time to reach the goal as crowd density increases. A ORCA-based simulator is not necessarily informative for an algorithm's performance, meanwhile, SFM adds higher levels of complexity to the testing scenarios. Navigation algorithms should seek training and testing scenarios that include a different level of complexity, heterogeneity, and cooperations from the human side, such as the one provided by the SFM.}
    \label{fig:densitystudy}
\end{figure}

\paragraph{Experimental Setup}

We consider an experimental setup in which a robot navigates within a crowd of simulated human agents in a circular workspace with a radius of $3 m$. All agents (the robot and humans), represented as disks of $0.3 m $ radius, are initially placed at random starting locations along the boundary of the workspace. Each agent is tasked with navigating from its starting location to the antipodal point on the circle. We consider two main environments\footnote{We use the implementation and parameters of \citet{chen19-crowdNav}, hosted at \url{https://github.com/vita-epfl/CrowdNav}. Note that in contrast to their default implementation, as soon as the agent reaches its goal, it is assigned a new target, randomly picked on the opposite center-symmetric side. We added this level of complexity, such that the simulated environment keeps remaining dynamic for the whole time the robot needs to reach his destination.} --an ORCA ~\citep{vandenberg} environment in which all human agents navigate by running the ORCA algorithm and a SFM environment~\citep{helbing2001} in which all human agents navigate by running the SFM algorithm. In parallel, the robot navigates by running one of the following algorithms: 
\begin{itemize}
    \item Simple Social Planner (SSP): A robot moving straight toward the goal and stopping when getting closer than $0.2 m$ to people. 
    \item Blind Planner (BP): A robot with no access to any sensory data.
    \item Optimal Reciprocal Collision Avoidance (ORCA) \citep{van2011rvo2}: A robot implementing the reciprocal collision avoidance algorithm given by \cite{van2008reciprocal}
    \item Socially Aware Reinforcement Learning (SARL)  \citep{chen17-sa-cadrl} : A robot that learns human-robot interactions with respect to their future states through an attentive pooling mechanism over pairwise interaction features
    \item Collision Avoidance Deep Reinforcement Learning (CADRL) \citep{chen17-cadrl} :  A robot that moves following a policy that tries to minimize the expected time to the goal subject to collision avoidance constraints 
    \item Relational Graph Learning (RGL) \citep{chen2019relational} : A robot that models the crowd as a spatial graph where agents are the nodes and edges represents human-robot interactions and uses a constant velocity model algorithm to predict agents position at the next time step  
    \item Model Predictive Relational Graph Learning (MPRGLP) \citep{chen2019relational} : A robot that models the crowd as a spatial graph where agents are the nodes and edges represents human-robot interactions and uses a model predictive algorithm to predict agents position at the next time step 
\end{itemize}

SARL, CADRL, RGL, and MPRGL have been trained using first imitation learning with collected experience from a demonstrator ORCA or SFM policy to initialize the model, and then RL to refine the policy. We use the subscript "SOCIAL" to differentiate the algorithms trained using a demonstrator SFM from the ones trained using demonstrator ORCA.
These algorithms have been trained for 10000 epochs using Adam Optimizer, a learning rate of $0.001$ and a discount factor of $0.9$.
The robot moves following holonomic kinematics, and the action space consists of $80$ discrete actions: $5$ speeds exponentially spaced between (0,$v_{pref}$] (where the preferred velocity is set to $1 m/s$) and $16$ headings spaced between $[0, 2\pi)$.

We consider eight different density levels, corresponding to scenarios involving 5, 7, 14, 21, 28, 35, 42, and 49 simulated human agents respectively. For each density level, we generate 100 random scenarios corresponding to distinct initial arrangements of agents along the boundary of the workspace. We examine the performance of the robot measured with respect to: a) average time to reach the goal and b) average number of collisions over the 100 trials. We then present a series of plots that make the case for a) the insufficiency of simulators commonly used for evaluating social navigation algorithms and b) the suboptimality of state-of-the-art social navigation frameworks.

\paragraph{Insufficiency of Simulation Environments}

In particular, Fig.~\ref{fig:densitystudy} reports the average time to reach the goal for SSP, ORCA, and BP as density increases in the ORCA and SFM environments, respectively. Surprisingly, in Fig.~\ref{fig:orca} we see that no matter how dense the crowd is, even a simple social planner such as SSP or BP significantly outperforms an ORCA-driven robot in an ORCA environment. ORCA agents are designed to be cooperative, allowing each other agent to take half of the responsibility of avoiding pairwise collisions. Besides, ORCA is only guaranteed not to collide and choose the maximum velocity in the next step, but it does not consider the shortest path overall. It lacks a mechanism to represent the general navigation task, while a naive planner such as SSP preserves this geometry information for path planning, achieving greater performance. 
For reference, in a more complex and heterogeneous scenario, such as the one represented by the SFM environment, the SSP underperforms compared to the other methods. It fails to reach the goal under a certain level of density is surpassed (Fig.~\ref{fig:SFM}), while the BP ends in a great number of collisions. This sort of behavior for basic planners is the one that is more likely to appear in a real world setting.

Our main takeaway from this example study is that testing \textit{only} on the ORCA simulator environment is not necessarily informative for an algorithm's performance. 
Additionally, the SFM adds higher levels of complexity to the testing scenarios. Hence, social navigation algorithms that can outperform these simple planners in such an environment, or an even more complex and heterogeneous setup, are most likely to show good performance in a real-world setting.

\paragraph{Suboptimality of State-of-the-art Approaches}





\begin{figure}[H]
     \begin{subfigure}{0.46\linewidth}
         \centering
         \includegraphics[width=\textwidth]{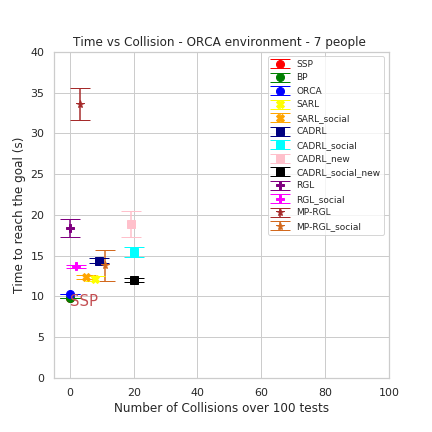}
         \caption{ Fully  cooperative  simulation  environment,  $ \approx 0.05$ people/$m^2$. The subscript \textit{social} indicates that the algorithms has been trained in a SFM environment.}
         \label{fig:boxplot_7}
     \end{subfigure}
          \begin{subfigure}{0.46\linewidth}
         \centering
         \includegraphics[width=\textwidth]{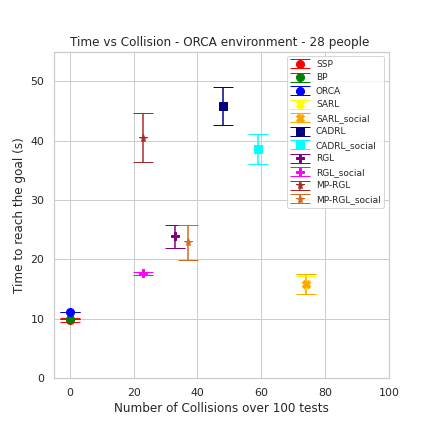}
         \caption{Fully  cooperative  simulation  environment,  $ \approx 0.19$ people/$m^2$. The subscript \textit{social} indicates that the algorithms has been trained in a SFM environment.}
         \label{fig:boxplot_28}
     \end{subfigure}\\
     \begin{subfigure}{0.46\linewidth}
         \centering
         \includegraphics[width=\textwidth]{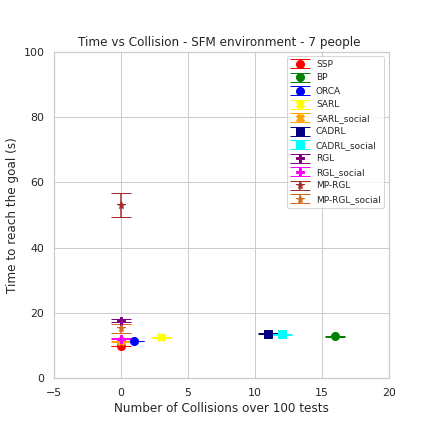}
         \caption{Fully  cooperative  simulation  environment,   $\approx 0.05$ people/$m^2$. The subscript \textit{social} indicates that the algorithms has been trained in a SFM environment.}
         \label{fig:boxplot_7_sfm}
     \end{subfigure}    
          \begin{subfigure}{0.46\linewidth}
         \centering
         \includegraphics[width=\textwidth]{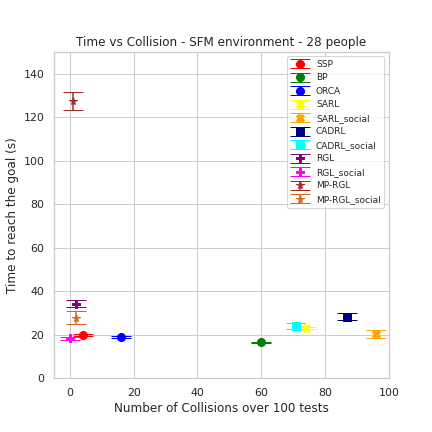}
         \caption{Fully  cooperative  simulation  environment,  {$\approx 0.19$ people/$m^2$}. The subscript \textit{social} indicates that the algorithms has been trained in a SFM environment.} 
         \label{fig:boxplot_28_sfm}
     \end{subfigure}
        \caption{Average  time  and  standard  error  to  reach the  goal  VS  number  of  collision  for  different  RL  algorithms  as  crowd  density  increases  on  ORCA (\ref{fig:boxplot_7},\ref{fig:boxplot_28}) and SFM (\ref{fig:boxplot_7_sfm},\ref{fig:boxplot_28_sfm}) environment. The SSP outperforms all state of the art RL methods, and is nearly identical to the best performing RL method.  Training in the SFM environment rather than in the ORCA environment strongly affects the policy’s performance. A robot trained in the ORCA environment results in a more conservative policy. On the other hand, a robot trained in the SFM environment moves  more  efficiently  (faster)  and  less  safely  (more collisions). Additionally, it shows  better  performances  when  tested  out-of-distribution,  represented  by the  ORCA  environment  in  this  specific  case.}
        \label{fig:collisionstudy}
\end{figure}

Fig.~\ref{fig:collisionstudy} reports the performance of all algorithms in the ORCA and SFM environments respectively for two density levels of interest: low density -- 0.05 $people/m^2$ (7 human agents) and high density -- 0.19 $people/m^2$ (28 human agents). Overall, we see the performance of the learning-based baselines does not transfer in domains falling outside of the test set distribution. 
Training in the SFM environment rather than in the ORCA environment strongly affects the policy's performance. A robot trained in the ORCA environment results in a more conservative policy. On the other hand, a robot trained in the SFM environment moves more efficiently (faster) and less safely (more collisions).
Additionally, an algorithm trained in a more complex and heterogeneous environment, such as the one represented by the SFM, shows better performances when tested out-of-distribution, represented by the ORCA environment in this specific case. These findings illustrate the importance of having a good high-level demonstrator, which, in our case, is given by the crowd's model. The algorithm chosen to represent the crowd's dynamics highly affects the performance of the learning algorithm. They also demonstrate that existing social navigation algorithms tend to struggle to outperform naively simple baselines in existing simulation environments, highlighting the lack of more realistic simulation suites and simulation benchmarking standards.




\paragraph{Behavioral Model Assumptions}

Another limitation, as pointed out by \citet{sim2real-crowds}, is that typical crowd simulators assume that all agents obey the same law, and all agents know the state of all other neighboring agents. However, individual behavior can affect the whole crowd's motion. 
Current crowd models fail to describe the relationships between human behaviors systematically and to enable heterogeneity. 
A central challenge for a crowd simulator is building models capable of integrating different crowd behaviors and interpreting how such actions affect the individuals’ movement under a unified mechanism representing different scenarios. 
Hence, one way to improve existing simulator suites would be to consider combining different modeling approaches that increase the crowd heterogeneity while modeling and representing complex individual behavior in different scenarios.
Additionally, it is possible to improve existing methods by varying the degrees of “attention” of the human agents. For example, a fraction of the agents could be considered inattentive, \exempli some people in crowds are looking at their phones. However, the agent should not be considered "fully" inattentive since the robot in the periphery could still trigger attentiveness.
A crowd simulator should also account for the agent's "flexibility", \idest how much humans are willing to compromise on their desired trajectory. 
People change their speed, desired directions, and they stop walking for several reasons. These rapid changes in the human pattern are not just due to the nearby people or obstacles, but to anything that caught their attention.





\subsubsection{Towards a Benchmarking Protocol}

Despite the limitations of existing simulation practices, simulation remains an important tool for social navigation research. Real-world studies are often costly, require extensive IRB reviews, and are challenging to design properly. Therefore, before proceeding to real-world testing procedures, it is crucial to acquire insights from repeated testing in simulation. Overall, as argued by \citet{oz_of_wizard}, methodologically rigorous human modeling is of high value role to human-robot interaction research. However, this necessitates addressing the human modeling issues raised above. Core research on the design of models for human locomotion and organization may offer significant insights into the design of models for human simulation~\citep{Warren2006,rio18}.

The complete validation of a social navigation framework requires extensive, statistically rigorous experimental testing in close proximity with humans in real-world environments. Existing literature in social navigation research is dominated by simulation studies with the limitations described in the previous sections, whereas experimental work is often lacking statistical significance (e.g., see Demos in Table \ref{tbl:littaxonomy}). In fact, much of the literature reports experimental demonstrations that inevitably capture a small, biased sample of rich behaviors that often emerge in real-world environments. However, even studies with more statistical rigor often suffer from design considerations. Specifically, existing lab studies often capture unrealistically simplistic or overly specific interactions with users. In contrast, the findings of field studies conducted in the wild may suffer from uncontrolled variables and the noise underlying open-form interaction with users. Furthermore, there is no consensus over objective or subjective measures for evaluating social navigation algorithms, neither a standardized repertoire of scenarios to test or platforms to experiment with.

Overall, these issues derive from the lack of a standardized experimental benchmark. Given the rich interest in the field, it is important as a community to converge towards a benchmark definition, clearly specifying all required experimental attributes. At the minimum, these should include: the exact definition of the task that the robot is performing; the types of experimental platform one should be using and the limitations of each (\exempli systems the implications of systems with nonholonomic kinematics); the number and role of humans in the experiment; the context and the role of the environment; the metrics we should be measuring performance with; the baselines we should be comparing against.

\subsubsection{Formal Verification}

It should be noted that the existence of established real-world benchmarks is not sufficient on its own to guarantee desirable performance. Any benchmark is still inevitably at best capturing a small sample of representative real-world interactions. Further, experiments are costly in terms of time and capital, prohibiting the possibility of exhaustive real-world testing. However, navigation in close proximity with humans is a safety-critical application and it is important that systems deployed in real-world settings are provably fail-safe. This highlights the important role that simulation still plays and the need for addressing the issues raised in Sec.~\ref{sec:experiments}. It also motivates the possible employment of methods for formal verification as practices to ensure robust performance. This is especially important considering the poor performance of existing algorithms on out-of-distribution test cases. While there is a wide body of work on the formal verification of robotic systems, including work specifically on collision avoidance~\citep{Mitsch17}, relatively limited attention has been placed to the more complex problem of verifying navigation systems operating in pedestrian spaces~\citep{Junges18}. Given the richness of pedestrian domains and the importance of delivering safe performance even under the highly likely emergence of system failures, more research is necessary to ensure graceful operation in realistic settings.

\section{Conclusion}\label{sec:conclusion}

We presented an overview of the field of social robot navigation with a survey spanning three decades of exciting developments. Acknowledging the important progress to date, we presented our perspective on a series of core challenges that prohibit mobile robots from seamlessly navigating in human populated environments today. We classified these challenges into three broad categories: planning, behavior design, and evaluation. 

First, we highlighted the importance of modeling interaction for social navigation research. We argued for the tight coupling between prediction and planning and traced its benefits in the literature. We also shared our thoughts on fundamental challenges underlying the problem of planning robot motion in crowded human environments. Specifically, planning in such spaces is NP-hard, and the non-convexity of balancing standard efficiency/safety tradeoffs prohibit guaranteeing practical performance in real-world applications. Additional complications arise from the difficulty of modeling human agents, and from the sensitivity to different types of contexts.

In terms of behavior design, we called for a need to organize and establish principled knowledge that governs social navigation behavior. We focused on spatial and motion behavior of pedestrians and mobile robots. And we reached the conclusion that literature in this domain are vast and varied, but no single theory stands out. It is generally accepted that considerations of certain behavior elements such as proxemics are important, but it is often unclear what is the accepted way to incorporate such elements. We argued that in order to establish such principled knowledge, both extensive testing in diverse large-scale environments and practicality testing by incorporating into social navigation models are needed for the proposed behavior models in the literature.

Finally, we pointed out issues with existing evaluation practices. We highlighted the shortcomings of commonly adopted simulation conventions. Surprisingly, we demonstrated that evaluating on the widely employed ORCA environment is not necessarily informative since a naive agent can leverage the cooperativeness of ORCA agents and outperform ORCA agents. Further, we saw that learning-based state-of-the-art approaches are extremely sensitive to out of distribution testing (equivalently, learning based methods do not generalize well). Motivated by these observations, we highlighted other strong assumptions used by existing simulation environments (homogeneity, fixed attention, and fixed flexibility) and argued for the importance of those parameters as design variables. Finally, we highlighted the lack of standardized benchmarks for social navigation research and argued for their importance as the community grows.

\subsection{Other Challenges}

While our categorization of open challenges captures a wide range of problems, the performance of social robot navigation frameworks is practically constrained by a number of broad challenges that were not directly within the scope of this review. 

\subsubsection{Robot Design}

Since the early days, roboticists have experimented with a wide variety of robot platforms, featuring a wide range of kinematic models, aesthetics, and anthropomorphic features.

In terms of kinematics, the prevalent paradigm involves wheeled mobile robots with a differential drive. While well-studied and universally available, such robots tend to produce motion that is not humanlike, often confusing humans. Recently, there have been notable alternatives, including the \emph{Ballbot} paradigm \citep{Nagarajan-hri09,Fankhauser_thesis,Lo19}, a robot resembling an inverted pendulum that is dynamically stabilizing on a single point of contact with the floor. Ballbots feature a humanlike form factor and agility that opens up the possibility of smooth merging in densely crowded environments. 

In terms of appearance, many works tend to ignore the aesthetics aspect, focusing on the development of customized functional systems or directly using existing robot platforms. Notably, RHINO \citep{rhino-robot} did not feature any anthropomorphic features but MINERVA \citep{minerva-robot}, the second-generation museum guide robot, featured a motorized face. Over the following years, many works employed a wide range of platforms with varying levels of humanlikeness, including anthropomorphic features such as a hat \citep{trautman-ijrr-2015} or a virtual face drawn on a screen \citep{Mavrogiannis19}. There is a long and extensive body of work in the HRI literature focusing on the tradeoffs of face design~\citep{DiSalvo02head,kalegina2018-robotface}. Further, it has been shown that the form factor itself is an important aspect of design when it comes to HRI applications~\citep{rae-takayama-mutlu}.

\subsubsection{Perception Challenges}

The planning and behavior prediction literature in social navigation often assumes that the robot is equipped with perfect on-board perception capabilities enabling precise robot localization and precise tracking of surrounding humans. However, outside of the lab, on-board perception will inevitably be limited and often prone to frequent failures. Thus, it is important that social robot navigation systems carefully account for the inherent perception uncertainty at the prediction and planning cycle. It is also important that they are designed with a principled mechanism for handling perception failures.

\subsubsection{Novelty Effects}

An important feature that is often not explicitly accounted for in the existing literature is the \emph{novelty effect}. While the widespread introduction of robot vacuum cleaners in households has contributed to the familiarization of humans to robots, it is expected---and in many cases anecdotally confirmed~\citep{trautman-ijrr-2015} -- that the spontaneous, unscripted encounter of humans with navigating robots in the wild features a great amount of unmodeled phenomena. The robot often becomes the pole of attraction, spiking the curiosity of bystanders who approach it, touch it and try to play with it, or challenge it (\exempli \citep{robot_abuse}). These types of interactions may complicate the robot's reaction and yield unpredictable and even unsafe responses. Finally, it is expected that these novelty effects will gradually dampen out as humans in a particular context become more familiar with a system. Therefore, it is important that the design follows principles of lifelong learning~\citep{THRUN199525}, accounting for the gradual adaptation of humans.

\subsection{Moving Forward}

The rich interest of the community in the area of social robot navigation and fast-paced neighboring fields, such as human behavior prediction~\citep{rudenko2019-predSurvey} and autonomous driving~\citep{autonomousdrivingsurvey}, are promising indicators for exciting developments over the next decade. To this end, we aspire that this survey will constitute an important asset for researchers as they plan their research agenda in the years to come.

\section{Acknowledgements}

This work was funded under grants from the Honda Research Institute USA, the National Science Foundation (NSF IIS-1734361), the National Institute on Disability, Independent Living, and Rehabilitation Research (NIDILRR 90DPGE0003), the U.S. Army Ground Vehicle Systems Center \& Software Engineering Institute at Carnegie Mellon University (PWP 6-652A4 DDCD GVSC), and the Air Force Office of Scientific Research (AFOSR FA2386-17-1-4660).

\bibliographystyle{ACM-Reference-Format}
{\bibliography{bibliography}}

\end{document}